%% file: main.tex
\RequirePackage[2020-02-02]{latexrelease}
\documentclass[dvipsnames]{article}

\usepackage[final]{neurips_2020}

\usepackage[utf8]{inputenc} 
\usepackage[T1]{fontenc}    
\usepackage[colorlinks=true,linkcolor=Blue,citecolor=ForestGreen]{hyperref}       
\usepackage{url}            

\usepackage{booktabs}       
\usepackage{amsfonts}       
\usepackage{nicefrac}       
\usepackage{microtype}      
\usepackage{times}
\usepackage{latexsym}
\usepackage{natbib}
\usepackage{amsmath}
\usepackage{xspace}
\usepackage{bbm}
\usepackage{graphicx}
\usepackage{color}
\usepackage{amsmath}
\usepackage{amssymb}
\usepackage{booktabs}
\usepackage{graphicx}
\usepackage{hyperref}
\usepackage{relsize}
\usepackage{latexsym}
\usepackage{paralist}
\usepackage{times}
\usepackage[printwatermark]{xwatermark}

\usepackage{url}
\usepackage{xcolor}
\usepackage{xspace}
\usepackage{diagbox}
\usepackage{multirow}
\usepackage{makecell}
\usepackage{bigdelim}
\usepackage{mathtools}

\usepackage{subcaption,dcolumn}
\newcolumntype{d}[1]{D..{#1}}

\newcommand\reddit{\textsc{Reddit}\xspace}
\newcommand\legal{\textsc{Legal}\xspace}
\newcommand\med{\textsc{Med}\xspace}
\newcommand\cs{\textsc{CS}\xspace}
\newcommand\realnews{\textsc{RealNews}\xspace}

\newcommand\gptthree{\textsc{GPT-3}\xspace}

\newcommand\demix{\textsc{DEMix}\xspace}

\newcommand\denselm{\textsc{Transformer-LM}\xspace}
\newcommand\denselmshort{\textsc{T-LM}\xspace}

\newcommand\delm{\textsc{ELMforest}\xspace}
\newcommand\dslm{\textsc{ELM}\xspace}
\newcommand\dslmfull{\textsc{expert LM}\xspace}
\newcommand\densetomod{\textsc{BTM}\xspace}
\newcommand\densetomodfull{\textsc{Branch-Train-Merge}\xspace}
\newcommand\random{\textsc{Random}\xspace}
\newcommand\unbalanced{\textsc{Unbalanced}\xspace}

\newcommand\gutenberg{\textsc{Gutenberg}\xspace}

\newcommand\cord{\textsc{CORD-19}\xspace}

\newcommand\aclpapers{\textsc{ACL Papers}\xspace}

\newcommand\oneb{\textsc{1B}\xspace}

\newcommand\contribfootnote[1]{%
  \begingroup
  \renewcommand\thefootnote{}\footnote{#1}%
  \addtocounter{footnote}{-1}%
  \endgroup
}

\title{Branch-Train-Merge: Embarrassingly Parallel Training of Expert Language Models}

\author{%
  Margaret Li$^{*\dagger\diamond}$ \\
  \And
  Suchin Gururangan$^{*\dagger\diamond}$ \\
  \And
  Tim Dettmers$^{\dagger}$ \\
  \AND
  Mike Lewis$^{\diamond}$ \\
  \And
  Tim Althoff$^{\dagger}$ \\
  \And
  Noah A.~Smith$^{\dagger\spadesuit}$ \\
  \And
  Luke Zettlemoyer$^{\dagger\diamond}$ \\
    \AND 
\textmd{$^{\dagger}$Paul G. Allen School of Computer Science \& Engineering, University of Washington }\\
$^{\spadesuit}$Allen Institute for AI \\
$^{\diamond}$Meta AI }
\begin{document}
\maketitle\vspace{-2em}
\contribfootnote{$^*$These authors contributed equally to this work. Correspondence to \{margsli, sg01\}@cs.washington.edu.}

\begin{abstract}
We present Branch-Train-Merge (\densetomod), a communication-efficient algorithm for embarrassingly parallel training of large language models (LLMs). 
We show it is possible to independently train subparts of a new class of LLMs on different subsets of the data, eliminating the massive multi-node synchronization currently required to train LLMs.  
\densetomod{} learns a set of independent \dslmfull{}s (\dslm{}s), each specialized to a different textual domain, such as scientific or legal text. These \dslm{}s can be added and removed to update data coverage, ensembled to generalize to new domains, or averaged to collapse back to a single LM for efficient inference. New \dslm{}s are  learned by \emph{branching} from (mixtures of) \dslm{}s in the current set, further \emph{training} the parameters on data for the new domain, and then \emph{merging} the resulting model back into the set for future use. 
Experiments show that \densetomod improves in- and out-of-domain perplexities as compared to GPT-style Transformer LMs, when controlling for training cost. Through extensive analysis, we show that these results are robust to different \dslm initialization schemes, but require expert domain specialization; LM ensembles with random data splits do not perform well.  We also present a study of scaling \densetomod into a new corpus of 64 domains (192B whitespace-separated tokens in total); the resulting LM (22.4B total parameters) performs as well as a Transformer LM trained with 2.5$\times$ more compute. These gains grow with the number of domains, suggesting more aggressive parallelism could be used to efficiently train larger models in future work.
\end{abstract}

\section{Introduction}
Training and inference in large language models (LMs) typically require access to supercomputers that can achieve the massive multi-node (e.g., GPU or TPU) synchronization required to compute model activations and gradients \citep{gpt3,opt,fedus2021switch,lepikhin2020gshard}. In this paper, we develop a new class of large LMs that is instead embarrassingly parallel: different parts of the model are independently trained on different subsets of the data, with no need for multi-node training or inference (Figure~\ref{fig:dense_vs_mod}). 

\begin{figure}
    \centering
    \includegraphics[width=\textwidth]{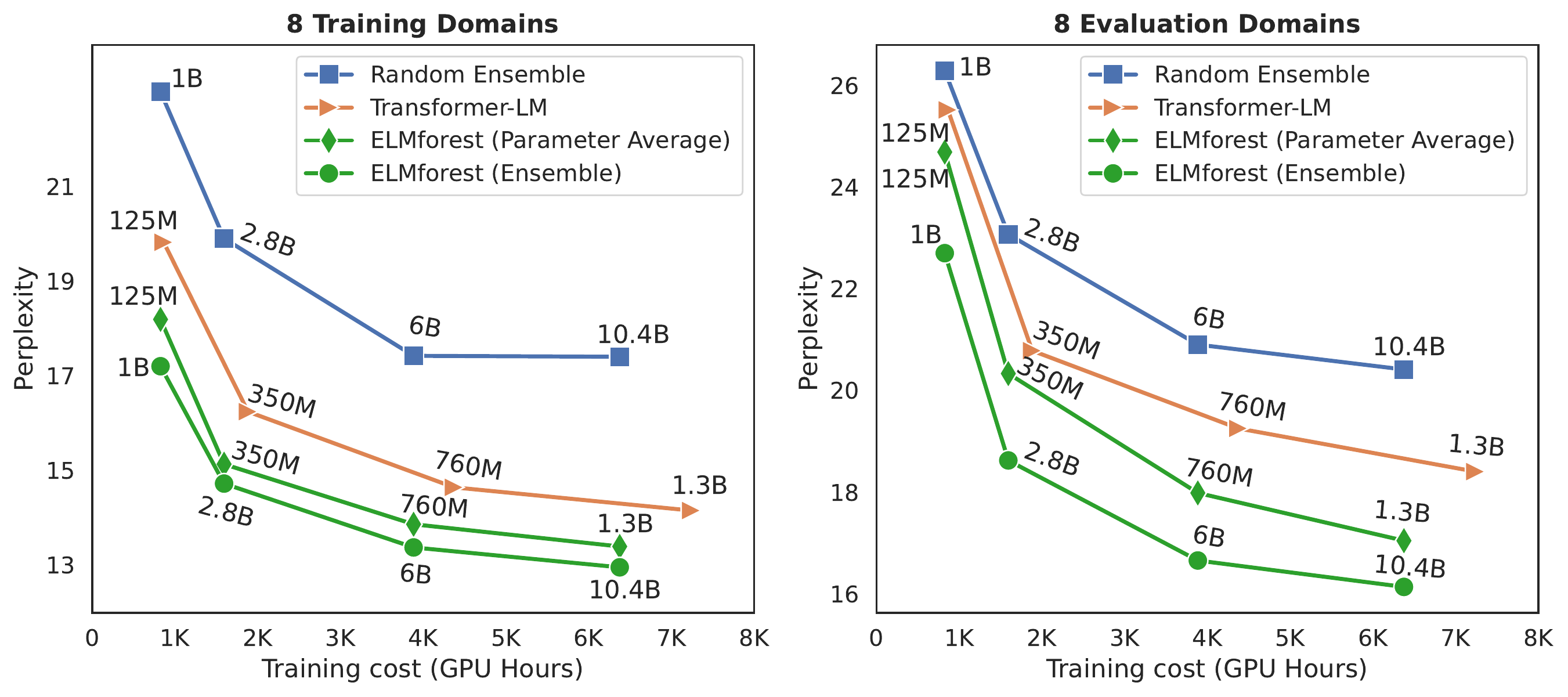}
    \caption{\textbf{\delm{}s outperform compute-matched \denselm and random-ensemble baselines across parameter scales (\S\ref{subsection:core_results})}. \delm parameter averaging closely approaches the performance of \delm ensembling at no additional cost to inference compared to the \denselm. \delm{} efficiency also grows as the model size increases due to substantially reduced GPU communication overhead relative to \denselm baselines.}
    \label{fig:compute_plot_fundamental}
\end{figure}
Our new \delm{}\footnote{\textbf{E}xpert \textbf{L}anguage \textbf{M}odels \textbf{For} \textbf{E}fficient \textbf{S}parse \textbf{T}raining} models consist of a set of \textbf{\dslmfull{}s} (\dslm{}s), each specialized to a distinct domain in the training corpus, e.g., scientific or legal text. 
The \dslm{}s are each independently functional LMs with {\em no shared parameters}, unlike previous domain mixture-of-experts models that only specialize the Transformer feedforward layers~\citep{demix}.
\dslm{}s can be added and removed to the model at any time to update data coverage, ensembled to generalize to new domains, or parameter averaged to collapse back to a single LM for more efficient inference.

We present the Branch-Train-Merge (\densetomod) algorithm for learning this set of specialized LMs. Our procedure repeatedly expands the \delm by adding one or more new
\dslm{}s completely in parallel. Each new \dslm{} in the \delm is first \emph{branched} by initializing a new LM with an average of the parameters of the most relevant LMs in the current set, then further \emph{trained} on new domain data with a standard cross-entropy loss, and finally \emph{merged} into the model by simply adding it to the current \delm (Figure \ref{fig:training_method} provides a schematic of this process). BTM is initialized with a single LM that is trained on heterogeneous data to establish strong shared representations for future domain specialization, a process that we explore extensively in our ablation analysis.

When evaluated in- and out-of-domain, \delm{}s trained with \densetomod outperform monolithic GPT-style transformer LMs (GPT-LMs) and a previous domain-specialized mixture-of-experts baseline (\demix; \citealt{demix}) across a range of computational budgets -- up to 1.3B parameters per \dslm{} trained for 7000 GPU-hours in aggregate (Figure~\ref{fig:compute_plot_fundamental}; \S \ref{subsection:core_results}). 
These gains are biggest for \delm ensembles, which use all of the model parameters, but also hold when we collapse the models by averaging parameters.  

We also perform detailed analysis to understand which aspects of \densetomod are most important for these gains. Ensembled \delm{}s outperform ensembling across random data splits, suggesting that domain specialization is a critical component to our approach  (\S\ref{subsection:random_domains}). 
We also show that performance is robust to a range of initializations, including the choice of the compute budget allocation (\S\ref{subsection:dense_ratios}) and data (\S \ref{subsection:dense_phase_corpus}) for training the initial LM. 
Our \delm{}s are also able to forget domains by removing the relevant \dslm{}, as long as they were not included in the initialization phase (\S\ref{subsection:dense_phase_corpus}).

Finally, we perform a preliminary scaling study on a training corpus with 192B whitespace-separated tokens (\S\ref{subsection:scaling_results}). Building on our findings, we use \densetomod to incrementally train a total of 64 experts which form a \delm. Our scaled \delm performs comparably with a 1.3B parameter \denselm{} trained with 2.5 times the total GPU hours. We find that benefits of \densetomod increase with the number of domains in the training corpus. 

These results provide compelling evidence for the promise of scaling large language models with many smaller, independently trained  \dslm{}s. We envision that this work lays the foundation for democratized model development at inclusive compute budgets --- that groups with different resource constraints and research interests may combine efforts to build open-sourced, community-authored large language models, comprised of continually-evolving repositories of \dslmfull{}s.

We release our code publicly.\footnote{ \url{https://www.github.com/hadasah/btm}}

Our contributions in this work are, concisely: 
\begin{itemize}
    \item \textbf{Branch-Train-Merge (\densetomod; \S\ref{sec:methods})}, consisting of an initial shared training phase, followed by parallelized training of many \dslm{}s  each specializing to one data domain. Inference with an \delm{}, either through ensembling or parameter averaging to produce a single LM,  outperforms compute-matched baselines at a variety of parameter scales (\S\ref{sec:core_results}).
    \item \textbf{Analysis (\S\ref{sec:analysis}).} We demonstrate that the improved performance of \delm{}s is not merely due to ensembling more parameters, and then study the performance tradeoffs of various training and inference decisions. \delm{}s trained with \densetomod robustly outperform compute- and parameter-matched approaches across many settings.
    \item \textbf{Incremental \densetomod training on 64 textual domains (\S\ref{sec:scaling_domains}).} We demonstrate increasing benefits of \densetomod training over a large baseline as we scale number of domains, up to at least 64 domains.

\end{itemize}

\section{\delm{}s}
\label{sec:model}

\begin{figure}
  \centering
    \includegraphics[scale=0.235]{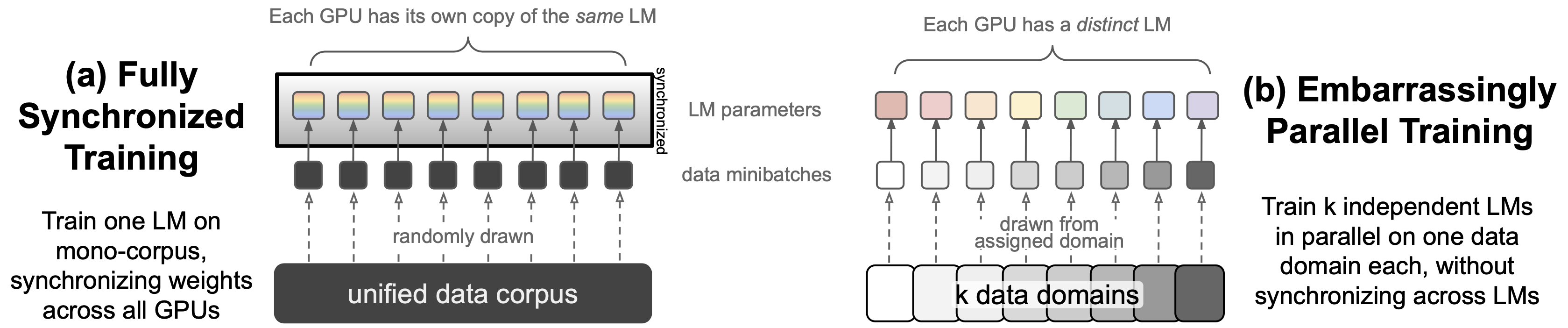}
  \caption{\textbf{Fully Synchronized vs.~Embarrassingly Parallel Training (\S\ref{sec:methods}).} 
  (a) In the fully synchronized data-parallel training of a \denselm, all parameters are synchronized across all GPUs. In large LMs, this synchronization incurs hefty cross-node communication costs. (b) In embarrassingly parallel training (our work), individual models are trained on each domain, with no parameter synchronization between those models, which eliminates cross-node communication costs between models. }
  
  \label{fig:dense_vs_mod}
\end{figure}

\delm{}s are designed to be embarrassingly parallel (Figure~\ref{fig:dense_vs_mod}) and rapidly customizable by completely isolating the influence of distinct data splits to different LMs, as defined in this section. 

\subsection{Model definition} We define an \textbf{\delm{}} to be a set of \dslmfull{}s (\dslm{}s), each independently trained to specialize to a different subset of a corpus. \dslm{}s are inspired by the experts in earlier MoE models \citep{jacobs1991adaptive}, but we specifically define ours to be \emph{domain} specialists and specialize the full LM instead of just model components. We follow \citealt{demix} and define domains by \emph{provenance}, or the source of the document (e.g., whether it is a legal document or computer science research paper), which yields simple and interpretable corpus segmentations, and is useful for identifying \dslm{}s in our experiments.\footnote{See \S\ref{sec:limitations} for a discussion on the possible limitations of this domain definition.} These can potentially be extended to multi-lingual, -modal, -task, or other types of data splits, but we leave these directions for future work. \dslm{}s remain independent at all stages of training and inference, enabling the functions described below.

\subsection{Adding and removing \dslm{}s}
\label{subsection:add_remove_experts}

We can modify the domain coverage of an \delm at any time by incorporating new \dslm{}s specialized to different domains or removing existing \dslm{}s in the set. These mechanisms stand in contrast to existing control techniques that have been introduced to steer LMs towards \citep{keskar2019ctrl,gururangan-etal-2020-dont,Dathathri2020PlugAP} or away \citep{welleck2019neural} from certain behaviors. Existing techniques tend to be expensive, require retraining the model with different objectives, or do not provide strong guarantees on how the LM may behave at test-time \citep{gehman2020realtoxicityprompts}. In contrast, \delm{}s 
allow for explicit inference-time application of constraints on the provenance of training data; they store all information from a domain in the associated \dslm, which is fully independent of all others. Removing any expert, even after fully training the model, guarantees the associated data will be fully ablated and never influence future model predictions. 

\subsection{Ensembling the \delm{}}
\label{subsection:mixing_experts}

\delm{}s support two inference modes, which trade off model efficiency and size for performance. In this first method, we ensemble the output probabilities of multiple \dslm{}s. This allows us to generalize to new text of unknown domain provenance. 
We use the \emph{cached prior} method proposed in \citet{demix}, summarized below.

Consider the probabilistic view of language modeling, where we estimate $p(X_t \mid \boldsymbol{x}_{<t})$. 
We introduce a domain variable, $D$, alongside each sequence. Then the next-step conditional distribution on the history $\boldsymbol{x}_{<t}$ is:
\begin{align}\small
    p(X_t \mid \boldsymbol{x}_{<t}) & \small{= \sum_{j=1}^n p(X_t \mid \boldsymbol{x}_{<t}, D = j) \cdot p(D = j \mid \boldsymbol{x}_{<t})}
\end{align}
We estimate a \textbf{domain posterior}, or a probability of a sequence belonging to each of the $k$ domains using Bayes' rule:

\begin{align}
    \small{p(D = j \mid \boldsymbol{x}_{<t})} &\small{= \frac{p(\boldsymbol{x}_{<t} \mid D = j) \cdot p(D = j)}{p(\boldsymbol{x}_{<t})} = \frac{p(\boldsymbol{x}_{<t} \mid D = j) \cdot p(D = j)}{\sum_{j'=1}^k p(\boldsymbol{x}_{<t} \mid D = j') \cdot p(D = j')}}
\end{align}

\dslm{}s  are used to compute the likelihood over contexts given a domain label. To compute the cached prior, we maintain an exponential moving average of posterior probabilities over domains, updated only at the end of each sequence block:
$p(D=j) = \sum_{i=1}^{N} \lambda^{i} \cdot p(D=j \mid x^{(i)}_{<T})$.
Following \citealt{demix}, we use $N=100$ sequences (of length $T=1024$ each) of development data, and set EMA decay $\lambda = 0.3$. We fix this prior at test time for each domain.

This inference procedure naively requires a forward pass through all \dslm{}s in the \delm{}, but we observe in practice that the domain posterior is sparse, even as the number of \dslm{}s increases, which suggests that top-$k$ selection of \dslmfull{}s can reduce inference time costs with negligible effects on performance. We quantify this sparsity and the effectiveness of using only the top-$k$ experts in our scaled-up experiments (\S\ref{subsection:inference_costs}).

\subsection{Averaging \dslm parameters} 
\label{subsection:averaging_experts}
As an alternative to ensembling, we can also use parameter averaging~\citep{https://doi.org/10.48550/arxiv.1803.05407,modelsoup, matena2021merging} to collapse the \delm{} into a single LM.  This operation keeps inference cost constant regardless of how many \dslm{}s are added to the set. When scaling the \delm to many domains, we also use \delm parameter averaging to initialize new experts, as we see in the next section. 
We experiment with several techniques to construct the average for a target domain in \S\ref{subsection:inference}, including a uniform average and using the best performing expert. We find that using the cached prior from \S\ref{subsection:mixing_experts} to define a weighted average LM is the strongest method. \delm{} weighted parameter averages outperform \denselm baselines at all model sizes studied, and continue to outperform \denselm{}s and approach \delm{} ensembling performance as we increase the number of domains (\S\ref{subsection:inference_costs}).

\section{\densetomodfull (\densetomod)}
\label{sec:methods}

\begin{figure}
  \centering
    \includegraphics[scale=0.22]{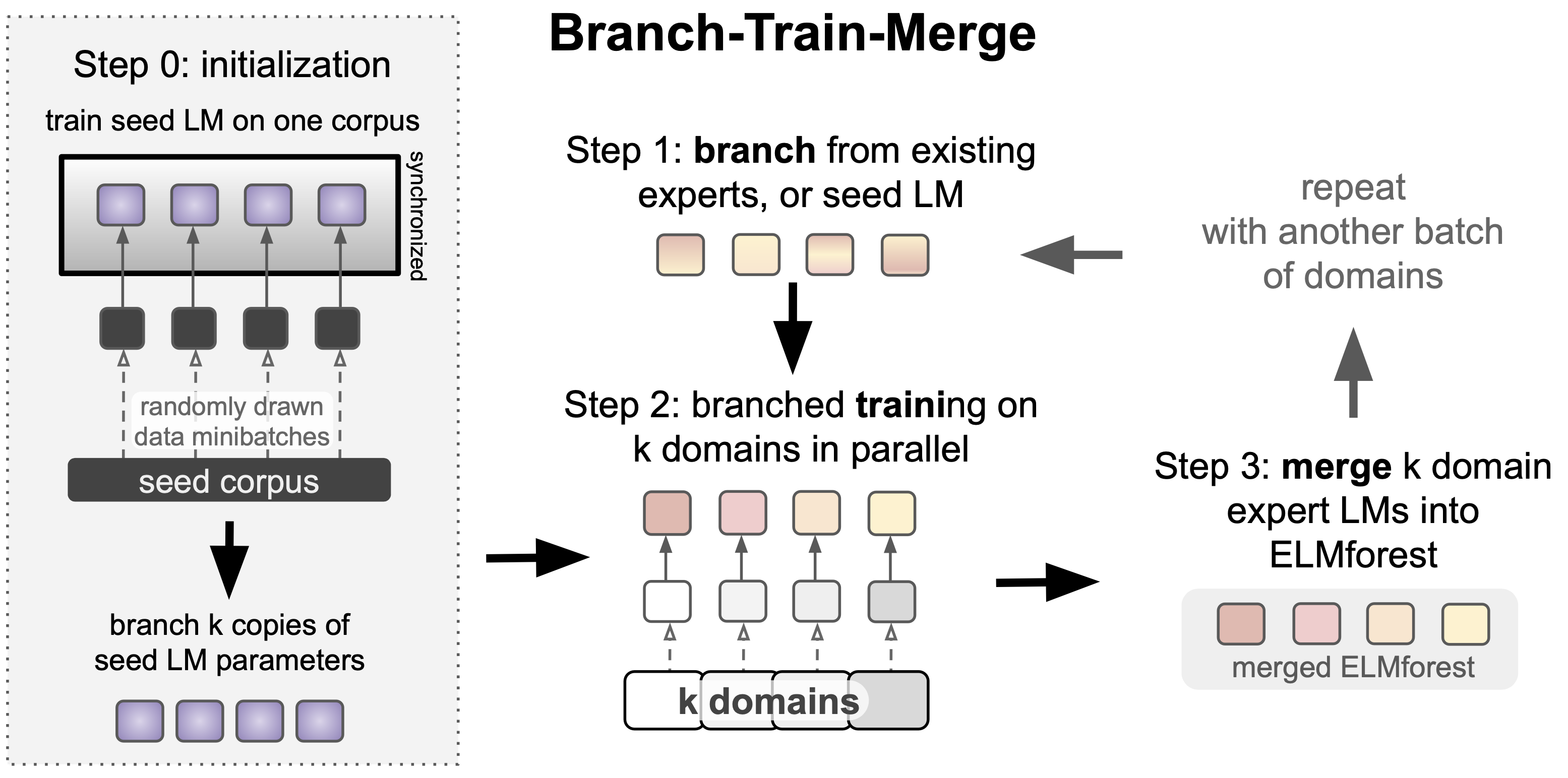}
  \caption{\textbf{\densetomod training process overview (\S\ref{sec:methods}).} 
Our \densetomod method is initialized with a seed training phase (Step 0), in which a single LM is trained on a corpus. The resulting parameters are branched, or copied, $k$ times (Step 1) and we continue training each LM on a unique assigned data domain, resulting in $k$ \dslm{}s (Step 2). Trained \dslm{}s are merged into the \delm{} (Step 3), and the process can be repeated from Step 1, initializing new \dslm{}s with parameter averages of existing \dslm{}s. After the seed phase (Step 0), \dslm{}s are fully disconnected, with no communication between them. This process is shown at $k=4$ for simplicity, but our experiments in \S\ref{sec:core_results}-\ref{sec:analysis} use $k=8$ domains and train for one iteration through \densetomod. Our experimental setup for \S\ref{sec:scaling_domains}, illustrated in Figure~\ref{fig:scaling_process}, trains on $k=64$ domains over 4 iterations of \densetomod.
}
  \label{fig:training_method}
\end{figure}

\densetomodfull training of \delm{} models is incremental and embarrassingly parallel (Figure~\ref{fig:dense_vs_mod}). \dslmfull{}s are trained completely independently in batches, starting from an LM trained on a shared, heterogeneous corpus. Figure~\ref{fig:training_method} summarizes the process, which we detail in this section.

\subsection{The Branch-Train-Merge Iteration}
\label{sec:btm3steps}

Each \densetomodfull{} iteration begins with an existing \delm{} $E = \{\theta_i\}_{i=1}^k$. Each \dslm{} $\theta_i$ represents a corresponding domain $d_i$ in the dataset of $k$ domains $D_E=\{d_1, \ldots , d_k\}$ currently modeled by $E$. In this section, we first describe the inductive case of $k>0$ and then describe how to train the initial model $\theta_0$. 

\paragraph{Step 1 (Branch): Sprouting a new \dslm{}}
 
The new \dslm{} parameters are a function of the current expert set $E$.
Given some vector of weights $w = \{w_1, w_2,...,w_k\}$ over the existing experts $\theta_1, \theta_2,...,\theta_k$, we can initialize the new expert with the weighted parameter average $\theta_{k+1}\leftarrow\sum_{i=0}^k w_i \theta_i$. 
We experiment with a few ways to set $w$, including initializing from only the nearest \dslm or -- our best performing approach -- a parameter average of \dslm{}s according to their domain posterior on the new data $d_{k+1}$ (\S\ref{subsection:inference}). 

\paragraph{Step 2 (Train): Growing the \dslm{}}
We train the new \dslm $\theta_{k+1}$ on data domain $d_{k+1}$ with the log likelihood objective. None of the existing \dslm{}s in $E$ are involved in the training of the new \dslm. We also refer to this step as \emph{branched training} later, to distinguish it from other training regimens we compare to.

\paragraph{Step 3 (Merge): Transplanting the \dslm{}}
We merge the new \dslm $\theta_{k+1}$, which now represents the domain $d_{k+1}$, into $E$ to create an updated set: $E' = E \cup \{\theta_{k+1}\}$ and $D_{E'} = D_E \cup \{d_{k+1}\}$. 

These operations are described for a single \dslm, but can be parallelized to add multiple \dslm{}s in a batch or iterated to add them asynchronously.

\subsection{Step 0 (Initialization): Seeding the \delm} 
\label{subsection:method_init}

On the first iteration of \densetomod, $E = \emptyset$; we have no \dslm{}s in the set to branch from. Instead of initializing the first \dslm{}s of the set randomly, we hypothesize that \dslm{} performance is boosted by branching from pretrained LM parameters, since multi-phase adaptive pretraining is an effective way to develop domain-specific language models \citep{gururangan-etal-2020-dont}, and parameter interpolation techniques work best with models that have a shared initialization \citep{https://doi.org/10.48550/arxiv.1803.05407,pmlr-v119-frankle20a, Wortsman_2022_CVPR, matena2021merging, modelsoup} . Specifically, we perform a \emph{seed phase}, training a \emph{seed LM} $\theta_{\textit{seed}}$ on some data corpus~$d_{\textit{seed}}$, which can be used to initialize the first batch of \dslm{}s in the set. In \S\ref{subsection:inference}, we find that the seed phase is important for enabling \delm parameter averaging (\S\ref{subsection:mixing_experts}). Ensembling \delm{}s trained with \densetomod results in comparable performance across a wide range of seed LM training settings (\S\ref{sec:analysis}). 

\subsection{Scale: Expanding the \delm } 
\label{subsection:method_repeat}

\delm{}s can be scaled incrementally by repeating the steps in \S\ref{sec:btm3steps} on batches of new domains. Incremental training allows new \dslm{}s to be initialized with parameter averages of existing LMs when branching. We hypothesize that this allows us to amortize the compute necessary to train new \dslm{}s as BTM proceeds; subsequent \dslm{}s can be trained for shorter periods of time by leveraging the pretraining of existing \dslm{}s in the set.

\section{Core Experiments and Results} 
\label{sec:core_results}

We first compare \densetomod training to compute-matched baselines, to carefully measure the efficiency gains. We use the simplest form of \densetomod training on a set of $k=8$ domains, consisting of one iteration through the Branch-Train-Merge cycle.
We compare inference through ensembling and different parameter averaging methods.

\subsection{Experimental Setup}
\label{subsection:core_experimental_setup}

\paragraph{Data} Following prior work, we use the data introduced by \citet{demix}, which consists of 8 training and 8 evaluation (primarily English-language) domains.\footnote{We refer to the novel domains from \citealt{demix} as evaluation domains in this paper.} These 16 domains cover a diverse span, from Web text and U.S court opinions for training to GitHub and COVID-19 research papers for evaluation. Details are listed in Appendix Table~\ref{tab:demix_corpus}.

\paragraph{Model hyperparameters} The model architecture is a randomly-initialized LM with the \gptthree \citep{gpt3} architecture implemented in Fairseq \citep{ott-etal-2019-fairseq} at the following model sizes, specified by parameter count: 125M (small), 350M (medium), 750M (large), 1.3B (xl). Following \citealt{gpt3}, we use the GPT-2 \citep{radford2019language} vocabulary of 50,264 BPE types, and train with 1,024-token sequences, across document boundaries. We prepend a beginning-of-document token to each document.

\paragraph{Compared Models}

\begin{itemize}
    \item \textbf{\denselm}: The first baseline is a standard Transformer LM, implemented with distributed data parallelism \citep{lidataparallel}. This is identical to the \textsc{dense} model from \citet{demix}, in which data from each domain is balanced.\footnote{We find, in line with \citet{demix}, that balancing data domains achieves better performance than without data balancing. Results comparing these two settings are in Appendix Table~\ref{tab:base_results_extended}.}
    \item \textbf{\demix}: We follow the training procedure outlined in \citet{demix}, where feedforward layers in the Transformer are trained to specialize as domain experts, but all other parameters are synchronized, as in the \denselm. \citet{demix} demonstrated that \demix LMs exhibit better domain specialization and generalization than other sparsely activated (e.g., MoE) models.
    \item \textbf{\delm}: We first conduct a seed phase to initialize the ensemble with LM parameters (\S\ref{subsection:method_init}), then conduct branched training on the \dslm{}s (\S\ref{sec:btm3steps}), all of which are initialized with the seed LM. Between the seed and branched phases, we continue training from the saved optimizer state.\footnote{We found in earlier experiments that resetting or reusing the optimizer state has little effect on performance; we reuse the optimizer state to fix the learning rate schedule across all experiments.} We ensemble the outputs of experts for all results with this model, using the procedure outlined in \S\ref{subsection:mixing_experts}.
\end{itemize}

These models are compute-matched, since computation is typically the limiting factor in model training. Like other sparse models \citep{fedus2021switch, lepikhin2020gshard, demix}, \delm{}s decouple compute and parameters; we can train many more parameters at the same computational cost as the equivalent \denselm{}. Total parameter counts are in Table \ref{tab:base_results}.

\paragraph{Training settings} To disentangle variations in GPU speed, we use \emph{number of updates} as our computational budget in these experiments. We choose the number of updates in our budget so that training completes in roughly 48 wall-clock hours: 80k, 32k, 24k, and 12k updates for the 125M, 350M, 750, 1.3B parameter per GPU models, respectively. We additionally report the average updates per second of each model, and present performance as a function of GPU hours, to illustrate efficiency improvements. We use 16, 32, 64, and 128 GPUs in parallel for the 125M, 350M, 750M, 1.3B parameter \denselm{} and \demix{} baselines, respectively. We also use these GPU budgets for the seed phase in the \delm{}. For branched training, we divide these GPU budgets equally among the \dslm{}s; for example, the 1.3B parameter per GPU \delm  uses 16 GPUs for each of the 8 \dslm{}s. For all models, we fix the learning rate at 0.0005 with a polynomial (linear) decay learning rate schedule and 8\% warmup, which we found to be optimal for most settings after a large grid search. For all experiments, we use a batch size of 16 for each GPU, with gradient accumulation of 32 steps, and train with \texttt{fp16}. We train on NVIDIA V100 32GB GPUs.

\subsection{Performance Comparisons
\label{subsection:core_results}}

\input{tables/base_results}

Our results are shown in Table \ref{tab:base_results}. 
Atbvfvhufckrjihhubfggkt these model scales, \delm{}s trained with the \densetomod procedure outperform both the sparsely trained \demix LM and the densely trained \denselm baselines. The improvements in performance we observe over \demix layers suggest that complete isolation of all LM parameters results in better specialization of domain experts. At the 125M parameter scale, we report the mean over 8 random seeds, as well as the standard deviation. 

\subsection{Efficiency Comparisons}
\label{subsection:efficiency}
\input{tables/train_speed}

Training \delm{}s requires less inter-GPU communication than \denselm or \demix models, since no synchronization occurs between GPUs assigned to different \dslm{}s. This results in higher updates per second and therefore shorter train times, as shown in Table \ref{tab:train_speed}.
Additionally, the embarrassingly parallel, fully disconnected nature of branched training provides flexibility in resource consumption; GPUs dedicated to different \dslm{}s may be online at different times, and \dslm{}s may even be trained serially on the same GPUs. Specifically, none of our branched training required more than 2 nodes of 8 GPUs simultaneously, so that we were able to conduct some \densetomod experiments on limited (e.g., academic-scale) resources. Our \denselm training experiments, on the other hand, consumed 16 nodes of 8 GPUs simultaneously. We observe this phenomenon throughout all experiments; empirically, \delm{} training jobs were scheduled and run more quickly, and with less preemption, than the \denselm and \demix training jobs at the same overall budget.

\subsection{\delm Parameter Average}
\label{subsection:inference}

While \delm substantially improves performance at lower training cost relative to the \denselm, it comes at the price of a larger model size and higher associated inference costs when ensembling. Here, we explore an alternative way to combine experts to improve generalization with no additional inference costs relative to the \denselm baseline: parameter averaging (\S\ref{subsection:averaging_experts}). Given some weight vector $w$ over $k$ \dslm{}s $\{\theta_i,...,\theta_k\}$, we define a single model such that all of its parameters are a weighted average of the \dslm{} parameters, according to $w$: $\theta = \sum_{i=0}^k w_i \theta_i$. For $w$, we consider a number of options:

\begin{itemize}
    \item \textbf{Uniform}: We set $w$ to be a uniform; i.e., $\frac{1}{k}$. This setting disregards the relevance of each \dslm{} to the target domain, assuming all \dslm{}s should contribute equally to the average.
    \item \textbf{Argmax} We set $w$ to be an indicator vector that corresponds to the maximum probability in the domain posterior. (\S\ref{subsection:mixing_experts}). This collapses the \delm into the estimated best-performing \dslm{} for the target dataset.
    \item \textbf{Posterior} We set $w$ to be the domain posterior (\S\ref{subsection:mixing_experts}), computed on the validation set.
\end{itemize}


We show our results on \delm parameter averaging in Table \ref{tab:averaging_results}. Using uniform weights underperforms all baselines, even \denselm{}s, highlighting the importance of domain relevance in ensembling and parameter averaging \dslm{}s. Using the argmax \dslm{} outperforms uniform averaging for small models, but not larger models. Weighting the average with the domain posterior outperforms all other techniques, and provides consistent performance improvements over \denselm{}s at no additional inference cost. 

The best parameter averaging performance does not reach the performance of ensembling. However, the lower inference costs and simplicity of deployment may make averaging the preferred inference technique for certain resource-constrained applications. We explore these computational tradeoffs further in \S\ref{subsection:inference_costs}. Due to its superior performance, we report results for \delm with ensembling, unless otherwise noted. 

Surprisingly, we observe poor performance of model averaging at the 125M scale regardless of the type of weight vector we use. We see later that the amount of compute allocated to the seed phase has a critical effect on the viability of \delm parameter averaging (\S\ref{subsection:dense_ratios}). With sufficient seed training, parameter averaging outperforms \denselm at all scales.

\input{tables/averaging_results}

\subsection{Summary}

\delm{}s trained with \densetomod demonstrate performance gains over update-matched \demix and \denselm{}s. Additionally, \densetomod is more efficient (higher updates per second) than \denselm or \demix training, due to a substantial reduction in cross-GPU communications for parameter synchronization. Collapsing the \delm{} into a single LM via parameter averaging results in consistent benefits over the \denselm at no additional inference costs, approaching the performance of \delm ensembling.

\section{Analysis}
\label{sec:analysis}
In the core results of \S\ref{sec:core_results}, we fixed the training setup to conduct a controlled comparison of \densetomod to baseline methods. Given the evidence supporting the improved performance of \delm{} trained with \densetomod, we now analyze the importance of various training and inference decisions. We study the effects of the increased parameter count in an \delm{}, of seed LM training data, as well as the compute budget dedicated to the seed phase, and also discuss the observed tradeoffs between performance, efficiency, and functionality.

\subsection{\delm performance is not simply the result of ensembling more parameters}\label{subsection:random_domains} 

\input{tables/random_domains}

To determine whether ensembling extra parameters is sufficient for the improvement our method provides, we compare multiple variations of LM ensembles:

\paragraph{Random Ensemble (seed init)} A set of LMs trained on random data splits, to assess the effect of removing the domain specialization of \dslm{}s. We first pool the training and development sets of our 8 train domains and divide into 8 random data splits, then execute the \densetomod procedure on those random splits, dedicating 50\% of training to the seed phase.\footnote{In Figure~\ref{fig:compute_plot_fundamental}, this setting is labeled \emph{Random Ensemble}.}

\paragraph{\delm (random init)} An \delm trained with \densetomod where all \dslm{}s are randomly initialized, to assess the effect of seed training. This is equivalent to setting the seed training compute budget to zero updates. We fix the random initialization across models.  

\paragraph{\delm (seed init)} The \delm setting of \S\ref{sec:core_results}, which follows the \densetomod training procedure on the 8 train domains, and splits the compute budget such that 50\% of the updates are dedicated to seed training and 50\% to branched \dslm training.

These results are in Table \ref{tab:random_domains}. \delm (random init) nearly matches \delm on training domains but performs poorly on evaluation domains. The random ensemble is consistently worse than both variants of \delm, showing that the performance improvement is not only due to ensembling or increased total model size. We speculate that the random ensemble is poor because its constituent models make correlated errors during evaluation \citep{gontijo-lopes2022no}.

\subsection{\delm performance robust to wide range of seed LM training compute allocations}
\label{subsection:dense_ratios}

\begin{figure}
        \centering
        \includegraphics[width=\textwidth]{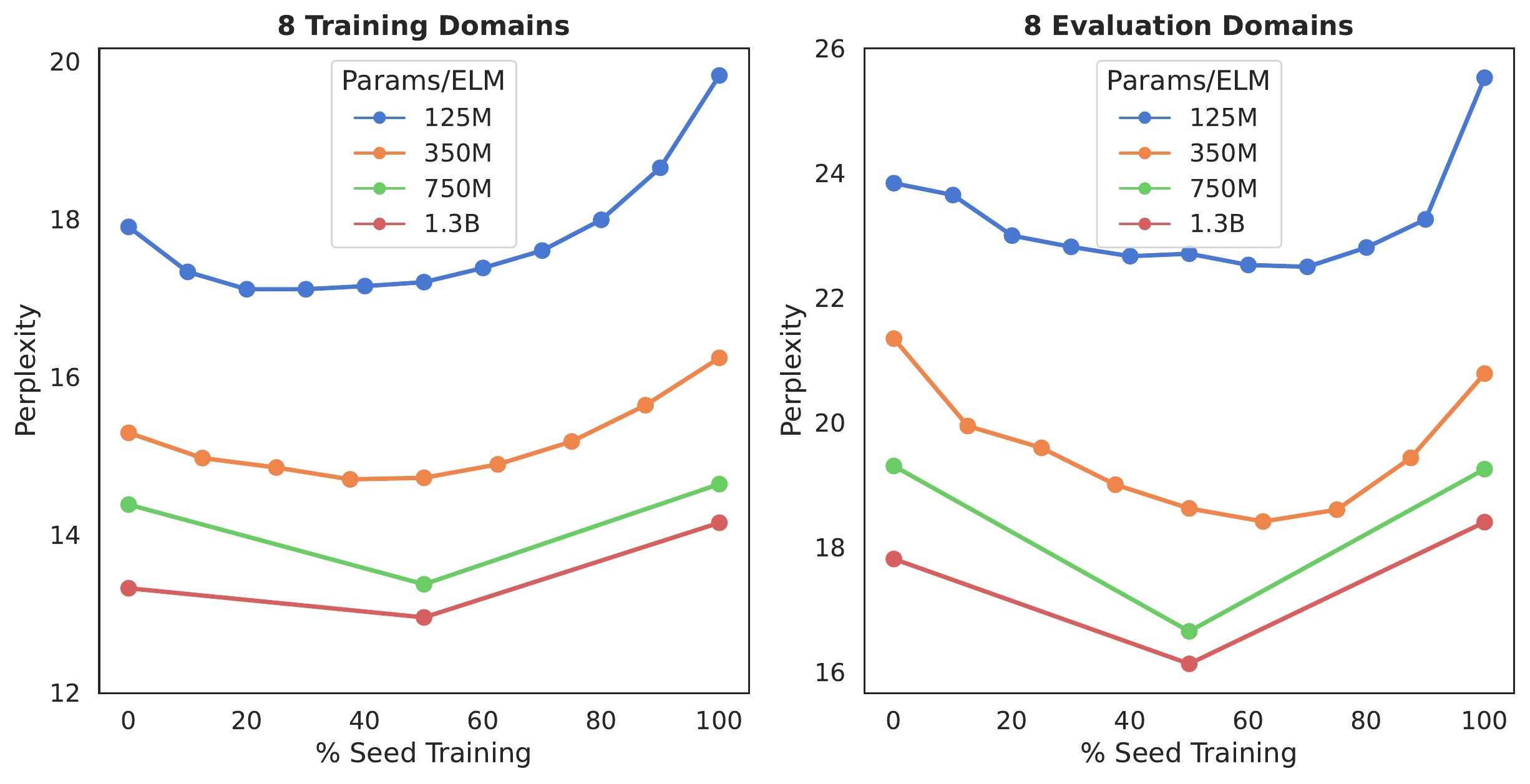}
        \caption{\textbf{\delm ensembling performance is robust to most seed training compute allocations (\S\ref{subsection:dense_ratios}).} Test perplexity averaged across the 8 training (left) or 8 evaluation (right) domains (from \S\ref{subsection:core_experimental_setup}) when varying the proportion of compute allocated to seed training, but keeping total compute budget fixed. All models use the same random initialization. The best training domain performance occurs when roughly 20--50\% of updates are seed training, whereas best evaluation domain performance occurs when roughly 50--80\% of updates are seed training. This supports the hypothesis that seed training is useful for generalization. However, performance varies little between 20--80\%, and almost all \densetomod settings perform better than both \denselm (100\% seed) and \delm-random init (0\% seed) training across model sizes.}
        \label{fig:dense_ratios}
    \end{figure}

The \delm (random init), which has no seed training, underperforms \delm (LM init) in \S\ref{subsection:dense_ratios}, indicating that the seed training phase is essential. On the other hand, \denselm, which is equivalent to 100\% seed training, also underperforms \delm (LM init) in \S\ref{sec:core_results}, which suggests the importance of branched \dslm{} training.  We now study the changes to performance when we vary the portion of the compute budget dedicated to seed training. We control for the total compute budget (across seed and branched training).

Our results, in Figure~\ref{fig:dense_ratios}, show that the optimal amount of seed training is about 40--60\% of the total budget. At both ends of the full range, performance deteriorates, approaching the \delm (random init) and \denselm performance (at 0\% and 100\% seed training, respectively).

However, as little as 10\% of seed training can be performed to result in strong gains over the \delm (random init) and \denselm. This suggests that the majority of \densetomod training may focus on branched \dslm training at dramatically reduced computational cost relative to the prevailing training of \denselm{}s, due to reduced GPU communication (\S\ref{subsection:efficiency}). The optimal share of compute to use towards each training phase likely depends on many factors, including the total compute budget. We leave more thorough study of this trend to future work.

\begin{figure}
    \centering
    \includegraphics[width=\textwidth]{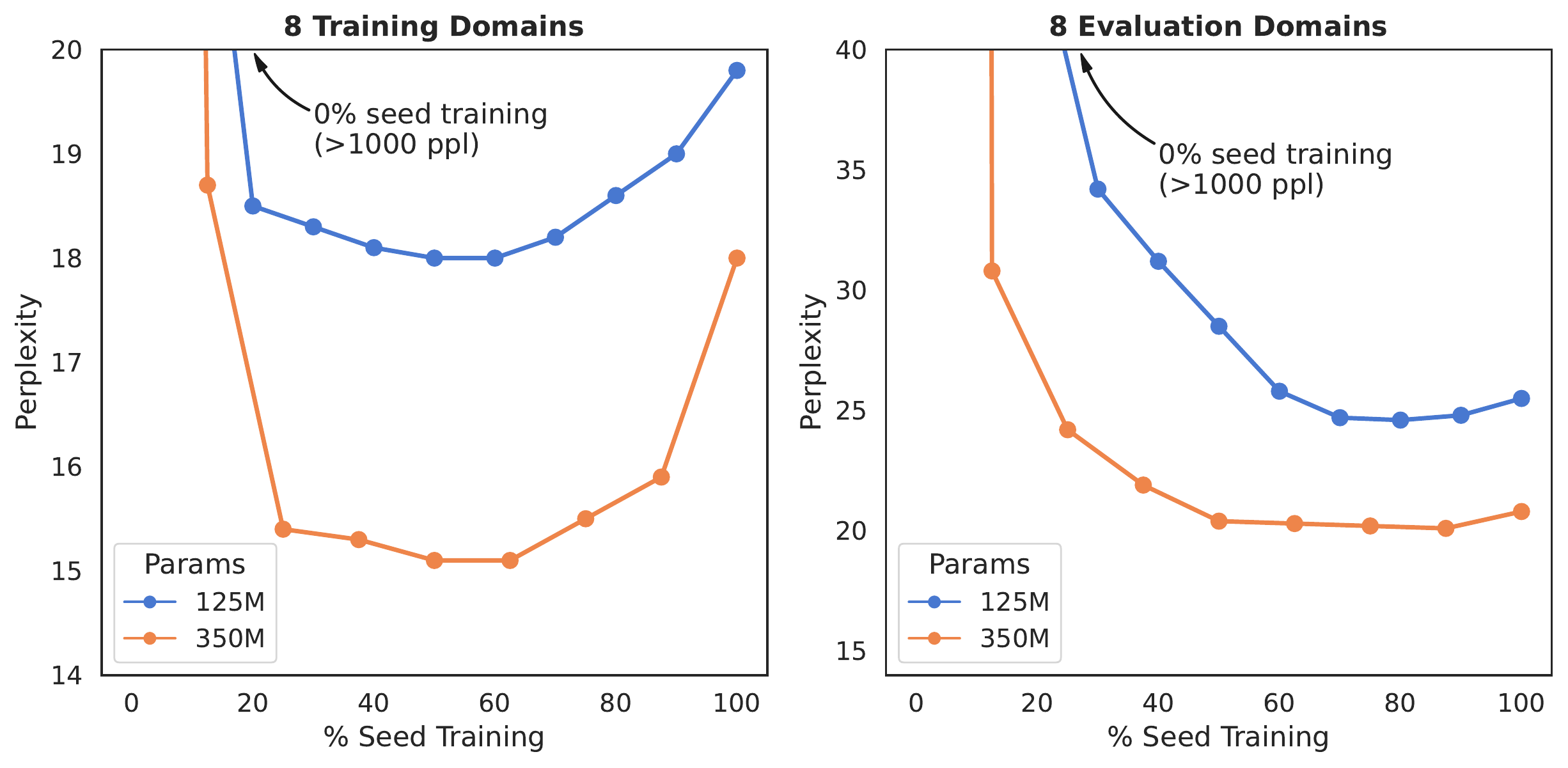}
    \caption{\textbf{The seed phase is vital to our ability to parameter average \dslm{}s (\S\ref{subsection:dense_ratios}).} Test perplexity averaged across the 8 training (left) and 8 evaluation (right) domains when averaging \delm{} with different seed training compute allocations for the 125M and 350M parameter LMs. All models use the same random initialization. Seed training is critical to reasonable averaging performance. Lowest perplexity is achieved at slightly higher seed phase training proportions than when using output ensembling; however, as with output ensembling, the variation in performance is small throughout a wide range of seed training phase lengths. For training domains, effective parameter averaging performance can be achieved with even less seed training.}
    \label{fig:averaging_across_ratios}
\end{figure}

\paragraph{\delm{} averaging performance strongest with about 60-70\% seed LM training}
Separate experiments, shown in Figure~\ref{fig:averaging_across_ratios}, suggest a strong effect of the seed phase on the viability of \delm{} averaging. We find that \delm{} averaging does not work with \dslm{}s trained from random initialization (i.e., with no seed phase), resulting in perplexities in the thousands. Since these \dslm{}s still share the same random initialization, we conclude that there is importance to seeding the \delm with a shared set of partially trained LM parameters. 

The averaging procedure is more robust to seed phase compute budget when evaluated on train domains. On evaluation domains, however, the smallest scale \delm{} does not achieve optimal performance until about 60\% or more updates are dedicated to seed training. This explains the poor performance of the 125M parameter scale \delm average on evaluation domains in Table~\ref{tab:averaging_results}. However, this optimal range shifts much lower, to about 40\%, for the next largest scale. Results for \delm parameter averaging at 50\% seed training for 350M, 750M, and 1.3B parameter scales do outperform \denselm baselines on both train and evaluation domains (Table~\ref{tab:averaging_results}).

\subsection{\delm performance is robust to the choice of seed training corpus \label{subsection:dense_phase_corpus}}

We compare the effects of using different training corpora for seed training in \densetomod. Here, we fix the compute budget allocations studied in \S\ref{subsection:dense_ratios} so that 50\% of updates are allocated to seed training and 50\% to branched training.
As seen in Table \ref{tab:dense_init}, our experiments using the most diverse corpora for seed training resulted in the best performance overall, but even seed training on JavaScript code alone yielded better results than the compute-matched \denselm baseline. 
This lack of sensitivity to seed phase training corpus suggests that initializing \dslm{}s with parameters of a model checkpoint is key -- perhaps regardless of the performance of that model on any target domain. However, the \delm (random init) models in Table \ref{tab:base_results}, which use identical random initialization, achieve worse performance, indicating the benefit of pretrained model parameters.

\input{tables/dense_init_results}

\paragraph{Domain forgetting through \dslm{} removal is mostly robust to seed training corpus}
\label{subsection:removing_experts_results}
\input{tables/expert_removal}
We introduced in \S\ref{subsection:add_remove_experts} the possibility of removing \dslm{}s to reduce the influence of specific training domains at inference time (e.g., those that contain stale or harmful text). Now, we evaluate the effectiveness of removing \dslm{}s in practice. In Table \ref{tab:expert_removal}, we show the performance of \delm{} ensembles on the training domains when using all \dslm{}s (from Table~\ref{tab:dense_init}), and compare to performance when removing the relevant \dslm{}. For example, to evaluate on \reddit, we keep active all \dslm{}s \emph{except} the \reddit \dslm. Removal of an \dslm from an \delm{} \emph{guarantees} that that domain will be forgotten, if the seed training corpus did not include that domain's data (\S\ref{subsection:add_remove_experts})\footnote{The actual effect of \dslm{} removal on model performance may depend on the overlap between the removed domain and other training domains. Future work may also investigate the effects of overlap and any required guarantees of domain disjunction.}. We find that performance does indeed degrade appreciably on \delm{}s when removing the \dslm, indicating that \delm{}s are capable of effectively forgetting a data domain without any gradient updates to the parameters. Even in the model with seed training on the 8 training domains, removal of a \dslm{} greatly degrades test perplexity in the corresponding domain. Taken together with the previous results, it may be possible to carefully curate the seed corpus (e.g., to avoid including harmful language) and domain data, to provide even stronger guarantees on the ability to forget unwanted domains via expert removal with minimal performance degradation.

\section{Incrementally Training an \delm on 64 Domains}
\label{sec:scaling_domains}

\begin{figure}
  \centering
    \includegraphics[scale=0.22]{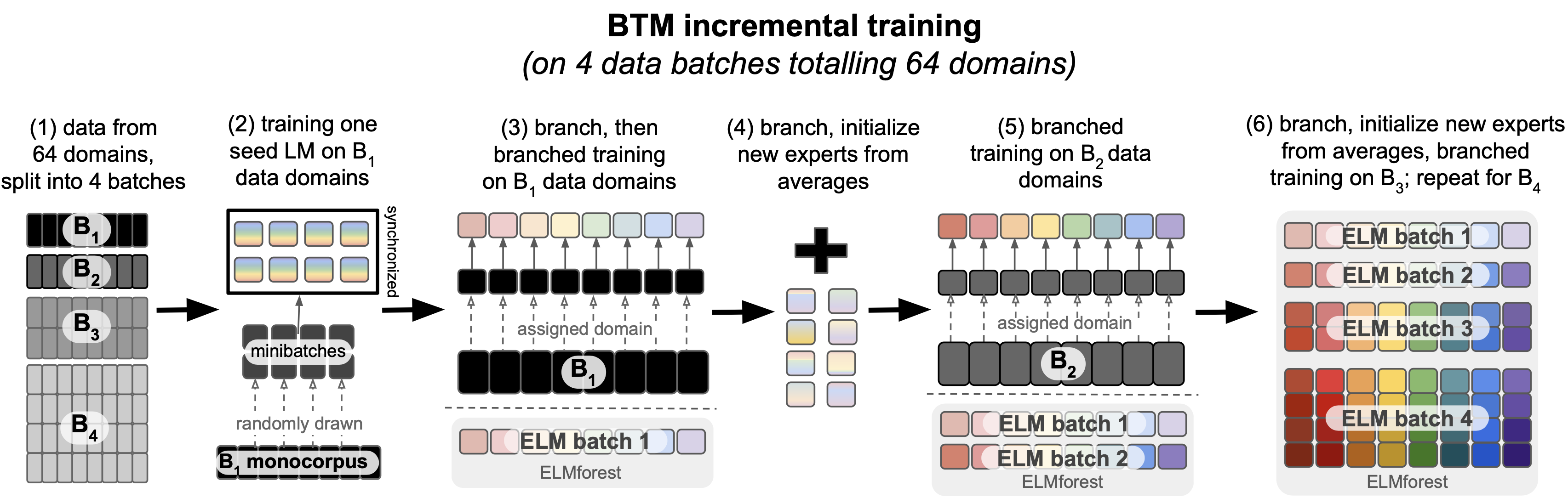}
  \caption{\textbf{\densetomod incremental training on 64 domains (\S\ref{sec:scaling_domains}).} 
We now iterate over the \densetomod steps (\S\ref{sec:methods}; Figure~\ref{fig:training_method}) to train incrementally on 4 data batches B$_{\text{1}}$--B$_{\text{4}}$, totalling 64 domains. After training the seed LM on the pooled B$_{\text{1}}$ domains, we train \dslm{}s on each B$_{\text{1}}$ domain independently, then initialize new \dslm{}s from weighted averages of existing expert parameters. These new \dslm{}s are each trained to specialize on a B$_{\text{2}}$ domain. We repeat this process for data domains in B$_{\text{3}}$, then B$_{\text{4}}$, initializing new \dslm{}s from the set of existing experts. After this full process, all 64 total \dslm{}s over 4 batches have been merged into the same \delm{} and may be ensembled or averaged together for inference.
}
  \label{fig:scaling_process}
\end{figure}

Next, using the best settings discovered in \S\ref{sec:core_results} and \S\ref{sec:analysis}, we repeat the BTM procedure to scale into an 80-domain corpus, which contains 64 training domains and 16 evaluation domains. Our training procedure is summarized in Figure~\ref{fig:scaling_process}. 
In \S\ref{subsection:scaling_results}, we demonstrate how the performance of the \delm scales relative to a large \denselm trained on the same set of domains from random initialization. We then examine the sparsity of the resulting \delm at inference time (\S\ref{subsection:sparsity}), and compare techniques to reduce the inference cost of larger \delm{}s (\S\ref{subsection:inference_costs}). 

\subsection{Experimental Setup}
\label{sec:scaled_experimental_setup}

\input{tables/scaling_dataset_alt}

\paragraph{Data: 80-domain Corpus}  Using provenance as our domain label, we curate a corpus of 64 training domains and 16 evaluation domains (Table \ref{tab:scaling_datasets}). Corpora were selected for their research-friendly data licensing, substantial size, and  content diversity, and are drawn from a collection of publicly available data repositories \citep[e.g.,][]{https://doi.org/10.48550/arxiv.2101.00027,lo-wang-2020-s2orc}. Full details on the data sources that make up these corpora can be found in Appendix Table \ref{tab:scaling_datasets_appendix} and \ref{tab:scaling_novel_datasets_appendix}.

\paragraph{Domain Batches} We first organize the 64 training domains into four ordered batches of increasing size, which form a curriculum that we denote B$_{\text{1}}$ to B$_{\text{4}}$. We maintain the training domains of \citet{demix} as the first batch (B$_{\text{1}}$), which allows us to leverage the best models from \S\ref{sec:core_results} and \S\ref{sec:analysis}, but organize the batches B$_{\text{2}}$-B$_{\text{4}}$ randomly. See Appendix Table~\ref{tab:scaling_datasets_appendix} for our batch assignments.

\paragraph{Model hyperparameters and training settings} We follow the same settings from \S\ref{subsection:core_experimental_setup}, but only compare models at the 350M (medium) and 1.3B (xl) scales.

\subsection{Compared Models}
\label{sec:compared_models}

The models in these experiments are intentionally \textbf{not} compute-matched, to demonstrate the efficiency gains of our approach over an expensive baseline. 

\paragraph{\denselm} Our baseline is a large, randomly-initialized 1.3B parameter transformer LM, trained for 6144 GPU hours (with 128 GPUs) on all 64 domains. We use the same training settings for this baseline as the \denselm in \S\ref{sec:core_results}.

\paragraph{\delm} We train on the 64 domains incrementally using 4 GPUs per ELM (see Figure \ref{fig:scaling_process} for a diagram of the overall process). For each batch of domains, we follow the basic procedure of BTM (\S\ref{sec:methods}):  branch a new set of experts, train them on the batch, and merge trained \dslm{}s back into the \delm. We start with \dslm{}s from \S\ref{sec:analysis} trained with 75\% seed training and 25\% branched training on the B$_{\text{1}}$ domains, which achieves the best evaluation-domain performance in our initial experiments (Figure \ref{fig:dense_ratios}). To train on B$_{\text{2}}$, we branch new \dslm{}s with weighted averages of B$_{\text{1}}$ experts, then train on B$_{\text{2}}$ for 40 GPU hours. For B$_{\text{3}}$, we branch new \dslm{}s again, with weighted averages of B$_{\text{1}}$ and B$_{\text{2}}$ experts, and train on B$_{\text{3}}$ for 40 GPU hours. Finally, we scale into B$_{\text{4}}$ by training for 20 GPU hours on each domain, initializing those \dslm{}s with the weighted average of \dslm{}s in B$_{\text{1}}$, B$_{\text{2}}$, and B$_{\text{3}}$.\footnote{The choice of compute budget at each training stage is an open question for future work. This may be informed by how much data existing \dslm{}s have already been exposed to and how much data exists in the group.} Our final \delm contains 64 \dslm{}s and is trained for a total of 2565 GPU hours.

\subsection{Results}
\label{subsection:scaling_results}

\begin{figure}
    \centering
    \includegraphics[width=\textwidth]{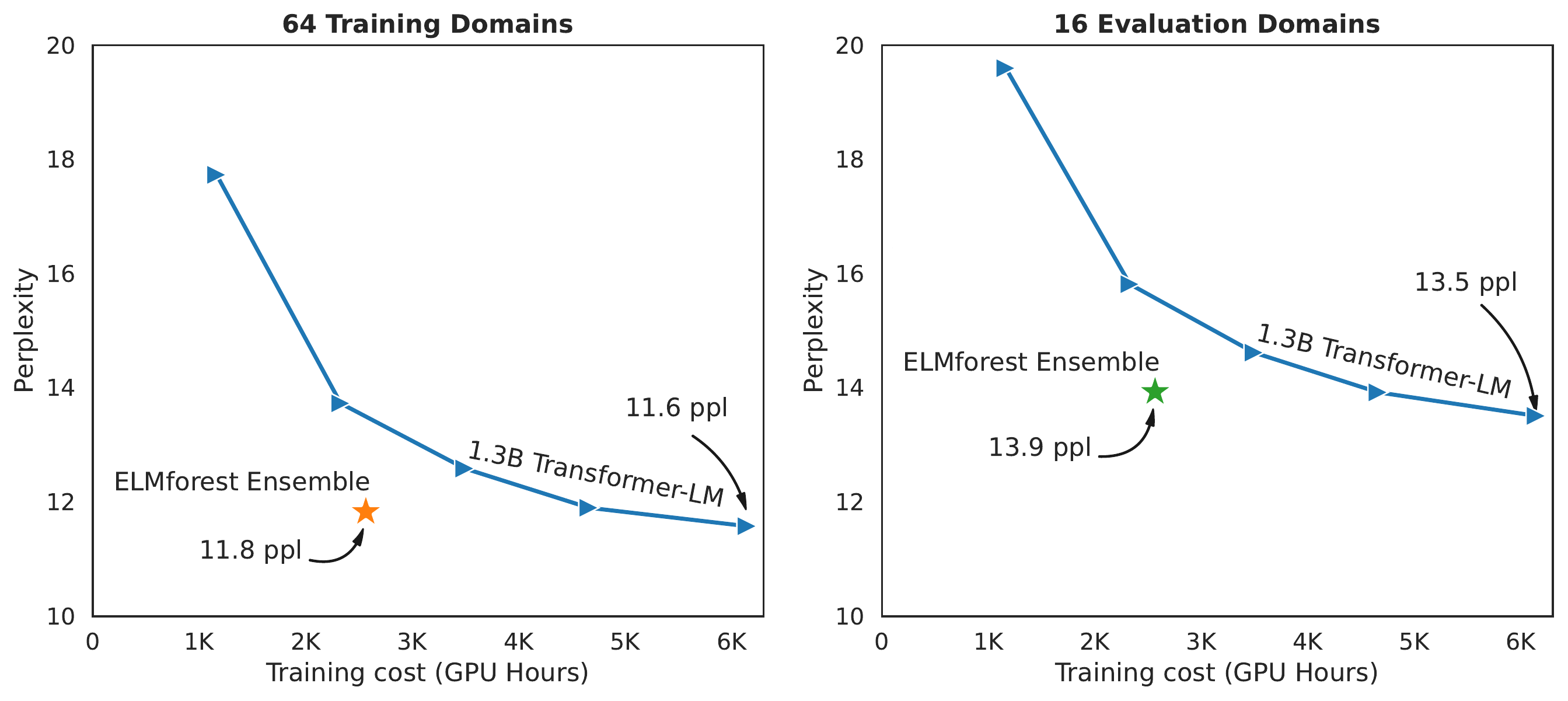}
    \caption{\textbf{\delm{} achieves performance comparable to a large \denselm baseline at a much smaller computational budget (\S\ref{subsection:scaling_results}).} Efficiency-performance trend for training (left) and evaluation (right) domains when \delm{}s  are incrementally trained on a 64 domain corpus, versus a large \denselm{} that is trained on 64 domains from scratch. \delm{}s achieve better perplexity at smaller compute budgets than \denselm{}s. \delm{}s perform comparably to the 1.3B \denselm{}, using only about 40\% of the \denselm{} total budget.}
    \label{fig:compute_plot_scaled}
\end{figure}
\hspace*{-1cm}

 Figure~\ref{fig:compute_plot_scaled} presents the performance of \delm and \denselm{} as a function of the total compute budget used for training on the 64 domains. We observe that  \delm with ensembling achieves perplexities on training and evaluation domains that are comparable to a large \denselm{}, despite using only 40\% of the total compute.\footnote{\delm parameter averaging approaches the ensemble in performance; see \S\ref{subsection:inference_costs} for details.} \delm effectively matches training domain performance of the large \denselm{} because the \denselm{} exhibits worse average performance on each individual training domain as it is exposed to more domains during training (see Appendix Figure \ref{fig:group_perplexities} for more details).\footnote{Previous work has shown that  phenomenon afflicts multilingual \denselm{}s as well, as more languages are added to the training corpus \citep{https://doi.org/10.48550/arxiv.2205.06266}.} On the other hand, we always observe monotonically non-worsening performance on each training domain with BTM, since previously trained \dslm{}s are unaffected when branching into new domains.

The order and composition of the domain batches likely affects the performance-efficency tradeoff curve for the \delm. We leave a careful analysis of these factors, and other optimal settings for scaling \delm{}s, to future work.

\subsection{Sparsity of the  64-expert domain posterior}
\label{subsection:sparsity}
\begin{figure}
      \centering
      \includegraphics[width=\textwidth]{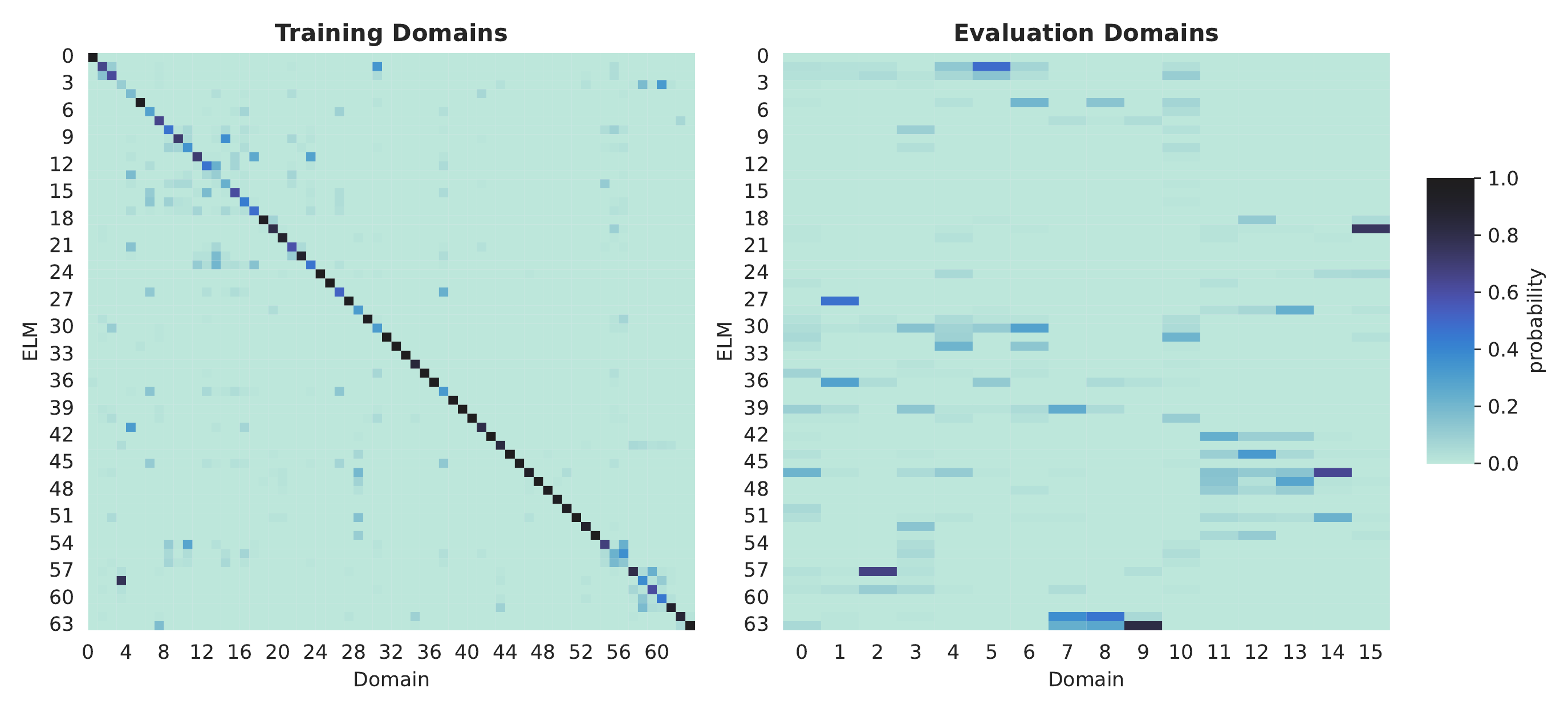}
      \caption{\textbf{Training and evaluation domain inference both use \dslm{}s sparsely (\S\ref{subsection:sparsity}).} Domain posterior visualization for 22.4B parameter \delm{} trained on 64 domains. \dslm{} activation is extremely sparse for both training and evaluation domains.} 
      \label{fig:heatmap}
  \end{figure}

A visualization of our resulting 64-expert \delm domain posteriors on held out validation data from the 80-domain corpus is displayed in Figure~\ref{fig:heatmap}. We observe sparsity in the domain posterior in both the training and evaluation domains. This sparsity suggests that only a few \dslm{}s need to be active at test time, a hypothesis we test next. We also display the top 3 \dslm{}s for a sample of evaluation domains in Table \ref{tab:topk_examples}. The most likely \dslm{}s are relevant to the evaluation domain. 

\input{tables/topk_examples}

\subsection{Reducing inference costs}
\label{subsection:inference_costs}

\begin{figure}
    \centering
    \includegraphics[scale=0.5]{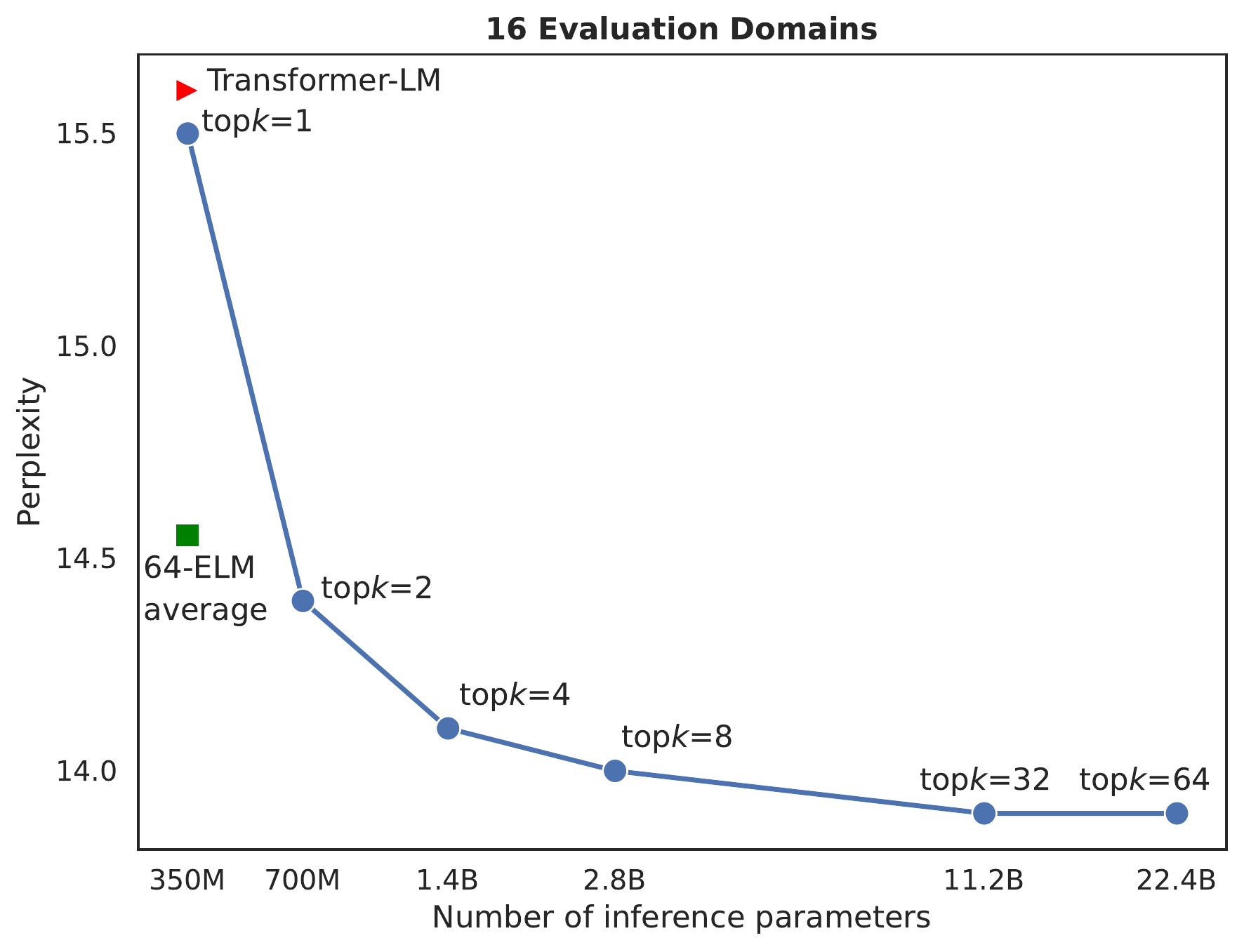}
    \caption{\textbf{Even using only the top-1 \dslm outperforms the \denselm baseline.} We vary the number of \dslm{}s active at inference time, and observe performance changes in the evaluation domains. In general, using just the top 4 or 8 \dslm{}s results in minimal performance loss relative to using 64 \dslm{}s at test time, owing to sparsity in the domain posterior. \dslm{} averaging results in a substantial performance improvement over \denselm{}s at the same inference cost; but larger gains are observed with larger top-$k$ values with output ensembling. }
    \label{fig:inference_plot}
\end{figure}

The sparsity of the domain posterior in the 64-expert \delm{} suggests that fewer \dslm{}s can be used at test time with minimal performance degradation. We see in Figure~\ref{fig:inference_plot} that this is indeed true; as few as the top-8 \dslm{}s can be ensembled together with negligible loss in performance relative to the 64-expert \delm{}. Even with just the top-1 \dslm{}, we see better performance for the training cost than a \denselm{}, at no additional inference cost.

On the other hand, weighted parameter averaging of all 64 \dslm{}s (with weights defined by the domain posterior) provides the best performance at the same cost of the original \denselm{}, but ensembling at least two \dslm{}s tends to outperform the \delm{}~average\footnote{This suggests that it may be possible to improve the algorithm to estimate weights for our parameter average. We leave this for future work.}. The runtime efficiency of averaged models may be desirable for resource-constrained applications, where using large LMs (i.e., those that cannot fit on a single GPU) is difficult. With averaging, inference costs stay constant regardless of the number of \dslm{}s in the \delm.

\subsection{Summary}

Our results demonstrate that a large collection of smaller \dslm{}s, each of which can be branched independently and on small compute budgets (e.g., at most four GPUs) with BTM, can perform comparably with a large \denselm{} trained with at least 2.5$\times$ the compute. While the costs of using all \dslm{}s in an ensemble at inference time is high, these costs can be reduced (with minimal performance degradation) to that of the equivalent \denselm{} through parameter averaging. Future work may perform closer investigations of best practices for scaling and coordinating \delm{}s to improve on the results we report.

\section{Related Work}

\paragraph{Sparse Language Models} Sparsely activated language models have been considered in a few forms \citep{pmlr-v119-evci20a,pmlr-v97-mostafa19a,dettmers-sparse-from-scratch}, but the Mixture-of-Experts (MoE) model is of particular note. Early versions \citep{jacobs1991adaptive} had independent feed-forward networks serving as experts. Recent MoE models \citep{shazeer2017outrageously} have been studied with token-based routing through backpropagation -- notably, by \citet{lepikhin2020gshard}, which appplies this concept to machine translation, and \citet{fedus2021switch}, which simplifies the architecture to activation of only one expert per layer. \citet{lewis2021base}, find an alternative approach to routing by formulating it as a linear assignment problem, and \citet{roller2021hash} use a fixed hash as the gating function.

Of this line of work, ours is most closely related to \citet{demix}. In that work, DEMix layers -- placed in the feedforward layers of the Transformer -- contain experts which specialize on specific domains. Routing at train time is determined only by the domain label, but all experts are activated at inference time and mixed according to weights estimated from a validation set.  Similarly, \citet{https://doi.org/10.48550/arxiv.2205.06266} develop a multilingual expert model with language-specific routing, and \citet{kudugunta-etal-2021-beyond-distillation} develop a multi-task expert model with task-specific routing.

\paragraph{Adapters} Previous work has also explored extending the capacity of a model with additional specialized parameters \citep[e.g., adapters;][]{pmlr-v97-houlsby19a, pfeiffer-etal-2020-mad, zaken2021bitfit}. However, unlike these existing approaches, our approach is significantly simplified, as our \dslm{}s each consist of an entire model which requires no additional parameters and no shared parameters. Future work may explore combining \dslm{}s with adapters to scale into smaller domains.

\paragraph{Ensembles} Ensemble methods are widely used in machine learning, for example in bagging, boosting, and stacking \citep{breiman1996bagging,freund1995boosting,wolpert1992stacked}. In a setting where training data is streamed, \citet{caccia2021on} define a \emph{growing ensemble}, in which new base models are trained sequentially on incoming batches. However, their growing ensemble, incrementally trained on the randomly created batches of their setting, underperforms non-incremental methods.

\paragraph{Parameter Averaging} Our averaging mechanism is inspired by the discovery that averaging many fine-tuned vision models improves out-of-domain generalization \citep{modelsoup,https://doi.org/10.48550/arxiv.1803.05407}. In \citealt{modelsoup}, the authors propose a greedy mechanism for averaging experts with uniform weights. Here, we find that uniform weighted averaging does not work for combining domain-specific models; instead we use a posterior weighted average, where the averaging weights are estimated based on the relevance of the model to the target domain. Our posterior weighted average is highly related to Bayesian model averaging techniques used in classic ensembling methods \citep{bma}. Model averaging has also been explored for federated learning \citep{mcmahan2017communication}, where different models are trained locally to fit privacy-sensitive data on different devices and merged. However, these works have found success averaging models trained from the same random initialization, which we do not find to hold in our setting. \citet{matena2021merging} compute a parameter average of models, estimating the optimal weights via an approximation of the Fisher information. Future work may explore these (and other) variations of weighted averages with \dslm{}s. 

\paragraph{Seed training} Our discovery of the importance of the seed training as a critical warm-up phase for BTM is in line with findings that parameter averaging only works when models share part of their optimization trajectory \citep{pmlr-v119-frankle20a, entezari2022the}. Future work may investigate what is learned in the seed phase that makes it so useful for \dslm specialization, regardless of the corpus used for seeding. Similar to seed training, \citet{https://doi.org/10.48550/arxiv.2112.14397} propose \emph{dense-to-sparse} gating, where mixture-of-experts routing mechanisms are gradually sparsified during the course of training.

\section{Limitations}
\label{sec:limitations}

\paragraph{The definition of a domain} The nature of domains in NLP is a matter of active research. Textual domains reflect language variation that stems from factors such as vocabulary differences \citep{blitzer-etal-2006-domain}, sociolinguistic \citep{biber_1988} or demographic \citep{Rickford1985EthnicityAA, blodgett-etal-2016-demographic} variables, community membership \citep{lucy2021characterizing}, end-tasks \citep{gururangan-etal-2020-dont}, or temporal shifts \citep{lazaridou2021mind, https://doi.org/10.48550/arxiv.2111.07408}. In this work, we follow \citet{demix} and define domains by \emph{provenance}, or the source of the document. Provenance labels yield simple and interpretable segmentations of a corpus, which are useful for identifying \dslm{}s in our experiments. However, other methods for discovering domains, including unsupervised techniques \citep{aharoni-goldberg-2020-unsupervised,chronopoulou-etal-2022-efficient}, may yield better expert assignments. We leave experimentation with other definitions of domain for future work.

\paragraph{Domain posterior data requirement} To calculate the domain posteriors used for our ensembling and parameter averaging weights, we assume access to a small additional sample of data to train the vector $w$. While it is easy to imagine that extra data may be available for most applications to estimate the posterior, future work may explore the possibility of eliminating this requirement.

\paragraph{Other distributed training baselines} Our \denselm{} baseline is implemented with distributed data-parallel. Model-parallel, fully sharded data-parallel, and other distributed training strategies \citep{https://doi.org/10.48550/arxiv.2112.10684} confer different scaling patterns that may change the conclusions that we report in this work. However, we expect that \densetomod will provide strong efficiency gains against these alternatives.

\paragraph{Harms of language models} BTM results in an LM whose test time behaviors can be controlled with much stronger guarantees after training due to the isolation of domains in \dslm{}s. However, \delm{}s  exposed to large datasets scraped from the Internet may contain toxic language (e.g., hatespeech) that are difficult to identify with coarse provenance domain labels, and nevertheless result in harmful output from the \dslm{}s \citep{gehman2020realtoxicityprompts}. Future work may explore recipes for training and deploying \delm{}s to better support user safety.

\section{Conclusion}

We introduce \densetomod training, a new algorithm to train an \delm{}, which contains many \dslmfull{}s that can be added and removed, ensembled, or parameter averaged at any time for efficient scaling and rapid customization. Our extensive experiments show that \delm{} ensembles trained with \densetomod outperform baselines at no additional training cost. Additionally, parameter averaged \delm{}s closely approach \delm{} ensemble performance while enabling substantially cheaper inference. These results provide compelling evidence for the promise of scaling large language models with many smaller, independently trained  \dslm{}s. We envision that this work lays the foundation for democratized model development at inclusive compute budgets --- that groups with different resource constraints and research interests may combine efforts to build open-sourced, community-authored large language models, comprised of continually-evolving repositories of \dslmfull{}s.

\acksection

This paper benefited from thoughtful feedback from a number of people: Ari Holtzman, Candace Ross, Colin Raffel, Gabriel Ilharco, Ishaan Gulrajani, Julian Michael, Mitchell Wortsman, Stephen Roller, Swabha Swayamdipta and William Fedus.

At UW, this work was partially supported by the Office of Naval Research under MURI grant N00014-18-1-2670.

\bibliography{custom,anthology}
\bibliographystyle{acl_natbib}

\vfill
\pagebreak
\appendix
\input{appendix}

\end{document}

%% file: tables/base_results.tex
\begin{table}[t]\small
\centering

\begin{subtable}[t]{0.48\textwidth}
\begin{tabular}{rccc}
\toprule
\multicolumn{4}{l}{\textbf{125M}} \\ 
\cmidrule{1-4}
& \denselmshort & \demix & \delm\\
& 125M & 512M & 1B \\
\cmidrule{2-4}
Train   & 19.9$_{0.23}$    & 18.2$_{0.82}$  & \bf 17.2$_{0.02}$ \\
Eval   & 25.2$_{0.18}$    & 23.4$_{0.54}$  & \bf 22.4$_{0.12}$ \\
All     & 22.5$_{0.14}$    & 20.8$_{0.63}$  & \bf 19.8$_{0.05}$ \\
\end{tabular}
\end{subtable}
\hfill
\begin{subtable}[t]{0.48\textwidth}
\begin{tabular}{rccc}
\toprule
\multicolumn{4}{l}{\bf 350M} \\ 
\cmidrule{1-4}
& \denselmshort & \demix &  \delm\\
& 350M & 1.8B & 2.8B \\
\cmidrule{2-4}
Train     &  16.3   & 15.0      &  \bf 14.7 \\
Eval     &  20.8   & 19.9      & \bf 18.6\\
All      & 18.5     & 17.5      &  \bf 16.7 \\
\end{tabular}
\end{subtable}

\begin{subtable}[t]{0.48\textwidth}
\begin{tabular}{rccc}
\toprule
\multicolumn{4}{l}{\bf 750M} \\ 
\cmidrule{1-4}
& \denselmshort & \demix & \delm\\
& 750M & 3.8B  & 6B \\
\cmidrule{2-4}
Train   & 14.7      & 13.5	    & \bf 13.4  \\
Eval   & 19.3      & 17.7      & \bf 16.7  \\
All     & 17.0      & 15.6      & \bf 15.0 \\
\bottomrule \\
\end{tabular}
\end{subtable}
\hfill
\begin{subtable}[t]{0.48\textwidth}
\begin{tabular}{rccc}
\toprule
\multicolumn{4}{l}{\textbf{1.3B}} \\ 
\cmidrule{1-4}
& \denselmshort & \demix & \delm\\
& 1.3B & 7B & 10.4B \\
\cmidrule{2-4}
Train   & 14.2      & 13.7      & \bf 13.0  \\
Eval   & 18.4      & 17.6      & \bf 16.3 \\
All     & 16.3      & 15.6      & \bf 14.6  \\
\bottomrule \\
\end{tabular}
\end{subtable}

\caption{\textbf{\delm{}s trained with \densetomod outperform all baselines across multiple model scales (\S\ref{subsection:core_results}).} Average test-set perplexity ($\downarrow$) for each model scale (125M, 350M, 750M, 1.3B parameters) across the 8 training, 8 evaluation, and all 16 domains described in \S\ref{subsection:core_experimental_setup}. Total parameters are shown for each model type at each scale. At 125M parameter per GPU scale, we show the mean and standard deviation of results over 8 random seeds. For \densetomod, we show results with 50\% of compute dedicated to the seed phase. \demix outperforms \denselm, abbreviated as \denselmshort. \delm{}s trained with \densetomod consistently achieve the lowest average test perplexity. }

\label{tab:base_results}
\end{table}

%% file: tables/train_speed.tex
\begin{table}[t]\small
\centering







\begin{tabular}{rccc}
\toprule
& \multicolumn{3}{c}{\bf Average updates per second, normalized $(\uparrow)$} \\
& fully synchronized & partially synchronized & \densetomod: embarrassingly parallel \\
& (\denselm) & (\demix) & (branched \dslm{}s) \\
\cmidrule{2-4}
\bf 125M & 1.00 & 1.01  & 1.05 \\
\bf 350M & 1.00 & 1.11  & 1.23 \\
\bf 750M & 1.00 & 1.01  & 1.27  \\
\bf 1.3B & 1.00 & 0.97  & 1.33 \\
\bottomrule \\
\end{tabular}

\caption{\textbf{\densetomod training is more efficient (\S\ref{subsection:efficiency}).} Average updates per second ($\uparrow$) for each setup and model size, normalized by the average updates per second during fully synchronized training of the \denselm. The embarrassingly parallel training used during the branched phase of \densetomod achieves higher updates per second than fully or partially synchronized training. The efficiency gains from embarrassingly parallel training become more substantial with larger model size -- and more nodes used in parallel. At 1.3B parameters per GPU, \densetomod's branched training provides a 33\% speedup over fully synchronized training. To better leverage this speedup, we experiment with dedicating a larger proportion of the total compute budget to branched \dslm training in \S\ref{subsection:dense_ratios}. These speed estimates may vary considerably with hardware and environment factors. }

\label{tab:train_speed}
\end{table}

%% file: tables/averaging_results.tex
\begin{table}[t]\small
\centering
\begin{tabular}{rcccc}
\toprule
& \multicolumn{4}{c}{\bf Train Domains PPL ($\downarrow$)}\\
& 125M & 350M  & 760M & 1.3B \\
\cmidrule{2-5}
\bf \denselm{}  & 19.9 & 16.3 & 14.7 & 14.2 \\
\bf \delm parameter average (uniform weights)  &  47.4 & 19.9 & 19.0 & 18.0  \\
\bf Argmax \dslm (one-hot posterior) & \bf 18.0  & 15.3 & 14.1 & 13.8 \\
\bf \delm parameter average (posterior weights) & \bf 18.0 & \bf 15.1 & \bf 13.9 & \bf 13.4 \\
\midrule
\bf \delm ensemble   & 17.2  & 14.7  & 13.4  & 13.0 \\



\end{tabular}
\begin{tabular}{rcccc}
\midrule
& \multicolumn{4}{c}{\bf Eval Domains PPL ($\downarrow$)}\\
& 125M & 350M  & 760M & 1.3B \\
\cmidrule{2-5}
\bf \denselm{}  & \bf 25.2 & 20.8 & 19.3 & 18.4 \\
\bf \delm parameter average (uniform weights)  &  31.0 & 22.4 & 20.8 & 19.5  \\
\bf Argmax \dslm (one-hot posterior) & 28.3  & 22.3 & 22.3 & 20.3 \\
\bf \delm parameter average (posterior weights) &  28.5 & \bf 20.3 & \bf 18.0 & \bf 17.0 \\
\midrule
\bf \delm ensemble    & 22.4  & 18.6  & 16.7  & 16.3 \\



\bottomrule \\
\end{tabular}
\caption{\textbf{\dslm{}s can be combined through parameter averaging (\S\ref{subsection:inference})}. Average test-set perplexity across the 8 training domains (top) and 8 evaluation domains (bottom), from the models in Table \ref{tab:base_results}, comparing techniques to collapse \delm{} into a single LM. Parameter averaging (with posterior weights) generally yields better average perplexities than \denselm{} at no additional inference cost, but underperforms \delm ensembling, which uses more effective parameters and is included for comparison as a lower bound. The relatively poor performance of \delm parameter averaging for the 125M parameter model is investigated further (and improved) in \S\ref{subsection:dense_ratios}.}

\label{tab:averaging_results}
\end{table}

%% file: tables/random_domains.tex
\begin{table}[t]
\small
\begin{subtable}[t]{0.49\textwidth}
\centering
\begin{tabular}{rccc}
\toprule
\multicolumn{4}{l}{\bf 125M} \\
\cmidrule{1-4}
& Random & \textsc{ELM} & \textsc{ELM} \\
& Ensemble & \textsc{forest} & \textsc{forest} \\ 
& (seed init) & (random init) & (seed init) \\
\cmidrule{2-4}
Train   & 23.0  & 18.2  & \bf 17.2 \\
Eval   & 26.0  & 23.4  & \bf 22.4 \\
All     & 24.7  & 20.8  & \bf 19.8 \\
\end{tabular}
\end{subtable}
\hfill
\begin{subtable}[t]{0.49\textwidth}
\centering
\flushright
\begin{tabular}{rccc}
\toprule
\multicolumn{4}{l}{\bf 350M} \\
\cmidrule{1-4}
& Random & \textsc{ELM} & \textsc{ELM} \\
& Ensemble & \textsc{forest} & \textsc{forest} \\ 
& (seed init) & (random init) & (seed init) \\
\cmidrule{2-4}
Train   & 19.9  & 15.3  & \bf 14.7 \\
Eval   & 23.1  & 21.3  & \bf 18.6 \\
All     & 21.5  & 18.3  & \bf 16.7 \\
\end{tabular}
\end{subtable}
\begin{subtable}[t]{0.49\textwidth}
\centering
\begin{tabular}{rccc}
\toprule
\multicolumn{4}{l}{\bf 750M} \\ 
\cmidrule{1-4}
& Random & \textsc{ELM} & \textsc{ELM} \\
& Ensemble & \textsc{forest} & \textsc{forest} \\ 
& (seed init) & (random init) & (seed init) \\
\cmidrule{2-4}
Train   & 17.4  & 14.4  & \bf 13.4 \\
Eval   & 20.9  & 19.3  & \bf 16.7 \\
All     & 19.2  & 16.9  & \bf 15.0 \\
\bottomrule \\
\end{tabular}
\end{subtable}
\hfill
\begin{subtable}[t]{0.49\textwidth}
\centering
\flushright
\begin{tabular}{rccc}
\toprule
\multicolumn{4}{l}{\bf 1.3B} \\
\cmidrule{1-4}
& Random & \textsc{ELM} & \textsc{ELM} \\
& Ensemble & \textsc{forest} & \textsc{forest} \\ 
& (seed init) & (random init) & (seed init) \\
\cmidrule{2-4}
Train   & 17.4  & 13.3  & \bf 13.0 \\
Eval   & 20.4  & 17.8  & \bf 16.3 \\
All     & 18.9  & 15.6  & \bf 14.6 \\
\bottomrule \\
\end{tabular}
\end{subtable}
\caption{\textbf{Domain expert ensemble outperforms random split ensemble (\S\ref{subsection:random_domains}).} Average test-set perplexity ($\downarrow$) for each model scale across the 8 training, 8 evaluation, and all 16 domains described in \S\ref{subsection:core_experimental_setup}. Training \delm experts with random data splits performs much worse than the proposed method of training \delm experts on domains of data defined by provenance. This suggests that ensembling with more parameters is not sufficient, but that domain specialization is a crucial component for the performance improvements we observe.}

\label{tab:random_domains}
\end{table}

%% file: tables/dense_init_results.tex
\begin{table}[t]\small
\centering


\begin{tabular}{rrccc}
\toprule
& & \multicolumn{3}{c}{\bf Average Test PPL ($\downarrow$)} \\
& & Train  & Evaluation & Overall \\
\cmidrule{3-5}
& \bf \denselm  & 19.8	&  25.5	& 22.7 \\
\cmidrule{3-5}

\parbox[t]{3pt}{\multirow{3}{*}{\rotatebox[origin=c]{90}{\bf {seed corpus}}}} 
& \bf 8 train domains   & \bf 17.2	& \bf 22.7  & \bf 20.0 \\
& \bf Wikipedia        & 17.7      &	23.2    &	20.5 \\
& \bf C4              & 17.9      &	23.5    &	20.7 \\
& \bf StackOverflow   &  18.4 &	24.6 &	21.5\\
& \bf JavaScript  & 19.2 &	24.9 & 22.0 \\
\bottomrule \\
\end{tabular}
\caption{\textbf{\delm{} ensembling performance is robust to seed training corpus (\S\ref{subsection:dense_phase_corpus}).} Test perplexity averages on the 8 training and the 8 novel test sets (from \S\ref{subsection:core_experimental_setup}), as well as averaged across all 16, with different training corpora used in seed LM training. All models are of the 125M parameters per GPU scale. All \delm{}s outperform the \denselm baseline. Notably, even training the seed LM on JavaScript results in lower perplexity than the \denselm, suggesting that \densetomod training is not overly sensitive to the selection of the seed training corpus. }
\label{tab:dense_init}
\end{table}

%% file: tables/expert_removal.tex
\begin{table}[t]\small
\centering

\begin{tabular}{lrrcc}
\toprule
&& &  \textbf{\thead{w/ \dslm{} \\  (Average Test PPL $\downarrow$)}} & \textbf{\thead{--\dslm  \\ ($\Delta$ PPL) }}\\
\midrule
&& \bf 8 train domains & 17.2 & (+9.4) \\
\cmidrule{4-5} 

\parbox[t]{1pt}{\multirow{4}{*}{\rotatebox[origin=c]{90}{\textbf{seed}}}} &
\parbox[t]{1pt}{\multirow{4}{*}{\rotatebox[origin=c]{90}{\textbf{corpus}}}} &
 \bf Wikipedia & 17.7 & (+11.9)\\
&& \bf C4       & 17.9 & (+11.8)\\
&& \bf StackOverflow  & 18.4 & (+12.7) \\
&& \bf JavaScript  & 19.2 & (+13.6) \\
\bottomrule \\
\end{tabular}


\caption{\textbf{The ability to reduce the influence of domains through \dslm removal is (mostly) robust to seed training corpus (\S\ref{subsection:dense_phase_corpus}).} We present the average test perplexity for the 8 train domains in \delm{}s where all \dslm{}s are active. We vary the seed training corpora. In parentheses, we show the \emph{increase} in perplexity when the \dslm trained to specialize on each domain is removed at inference time. Large perplexity increases are desired, as they suggest the ease of removing data from the \delm{}'s distribution after training has completed. This is a useful operation to reduce the influence of training domains that contain harmful text or become stale.}
\label{tab:expert_removal}
\end{table}

%% file: tables/scaling_dataset_alt.tex
\begin{table*}[t]
\centering
\tiny
\caption*{\textsc{80-domain Corpus: 192.3B whitespace-separated tokens}}
\resizebox{\textwidth}{!}{
\begin{tabular}{p{3cm}p{9cm}}

\toprule
\bf Category & \bf Domains   \\
\midrule 

\textsc{Semantic Scholar} (26.6\%) & Medicine (5.2\%),  Biology (4.7\%),  CS  (3.4\%),  Physics  (2.7\%),  Math  (2.3\%),  Unlabeled  (1.3\%),  Psychology  (1.2\%),  Chemistry  (1.0\%),  Economics  (0.8\%),  Engineering  (0.7\%),  CORD19 (0.6\%),  Material Science  (0.5\%),  Geology  (0.5\%),  Sociology  (0.5\%),  Business  (0.3\%),  Political Science  (0.2\%),  Geography  (0.2\%),  Environmental Science  (0.1\%),  History  (0.1\%),  Philosophy  (0.1\%), ACL (0.1\%),  Art  (0.05\%) \\
\midrule
\textsc{Github Code} (22.4\%) & JavaScript (3.7\%),  Java (3.5\%),  HTML (2.7\%),  C (2.5\%),  C++ (1.9\%),  Python (1.5\%),  C\# (1.2\%),  PHP (1.1\%),  Markdown (1.1\%),  Code Contests (1.0\%),  GO (1.0\%),  CSS (0.7\%),  Ruby (0.4\%) \\
\midrule
\textsc{Web Forums} (17.5\%) & Reddit Dialogues (13.0\%),  StackOverflow (1.7\%),  Twitter (0.9\%),  StackExchange (0.8\%),  HackerNews (0.4\%),  Gaming Subreddits (0.1\%),  Sports Subreddits (0.1\%) \\ 
\midrule
\textsc{Web Crawl} (16.0\%)& C4  (5.2\%),  RealNews (5.2\%),  OpenWebText (3.4\%),  Wikipedia (en) (1.3\%),  WMT News Crawl 2021 (0.5\%),  1B Words Corpus (0.4\%) \\
\midrule
\textsc{Books} (5.8\%) & Stories (3.8\%),  Gutenberg Books (1.6\%),  BookCorpus (0.4\%) \\
\midrule
\textsc{Legal Text} (5.5\%) & Legal Case Law (5.5\%),  Supreme Court Opinions (HTML) (0.1\%) \\
\midrule 
\textsc{Reviews} (5.0\%) & Books Reviews (2.1\%),  Amazon Reviews (1.1\%),  Electronics Reviews (0.5\%),  Clothing, Shoes and Jewelry Reviews (0.5\%),   Home and Kitchen Reviews (0.4\%), Yelp Reviews (0.3\%),  Sports and Outdoors Reviews (0.3\%),  Movies and TV Reviews (0.3\%) \\
\midrule
\textsc{Other} (1.3\%) & DM Mathematics (0.8\%),  OpenSubtitles  (0.4\%),  USPTO  (0.1\%) \\
\midrule


\textsc{Evaluation Domains \newline (Test Only)} & Enron, \#COVID-19 Tweets, IMDB, TOEFL exams, Congressional bills, Legal Contracts, /r/cscareerquestions, /r/india, /r/hiphopheads, Irish Parliamentary Speeches, SQL, Rust, Perl , TeX, FORTRAN, Breaking News \\

\bottomrule
\end{tabular}

}
\caption{\textbf{Overview of the 80-domain corpus (\S\ref{sec:scaled_experimental_setup}).} The 80 domains that make up the multi-domain corpus we train and evaluate on, presented here in 8 descriptive categories for ease of inspection. For each of the 64 training domains, we include the percentage of the total number of tokens (in the entire corpus) comprising that domain. At the bottom, we include the 16 evaluation domains. All domains additionally include 1M tokens for validation and test data each. We include full details of each corpus in Appendix Table \ref{tab:scaling_datasets_appendix} and \ref{tab:scaling_novel_datasets_appendix}.}
\label{tab:scaling_datasets}
\end{table*}

%% file: tables/topk_examples.tex



\begin{table}[t]
\centering
\small
\begin{tabular}{rrrr}
\toprule
\textbf{Evaluation Domain}  & \textbf{Top-1} & \textbf{Top-2} & \textbf{Top-3} \\
\midrule
\emph{Covid Tweets} & Twitter (0.48) & 2021 Newscrawl (0.29) & HackerNews (0.05) \\
\emph{TeX} & stackexchange (0.64) & Markdown (0.22) & HTML (0.05) \\
\emph{Congressional Bills} & SCOTUS Opinions (0.21) & OpenWebText (0.13) & Stackexchange (0.11) \\
\emph{IMDB} & Movie \& TV Reviews (0.66) & Book Reviews (0.11) & RealNews (0.05) \\
\emph{Legal Contracts} & C4 (0.30) & Legal (0.21) &  SCOTUS Opinions (0.14) \\






\bottomrule \\
\end{tabular}

\caption{\textbf{Top-3 \dslm{}s (with probabilities in parentheses) under the domain posterior for a sample of evaluation domains (\S\ref{subsection:sparsity}).} The most likely \dslm{}s under the domain posterior are usually highly relevant to the evaluation domain by inspection.}
\label{tab:topk_examples}
\end{table}

%% file: appendix.tex
\section{Appendix}



\input{tables/datasets}
\input{tables/scaling_datasets_appendix}

\input{tables/scaling_novel_datasets_appendix}
\input{tables/anonymization}

\input{tables/base_results_extended}

\begin{figure}
    \centering
    \includegraphics[scale=0.5]{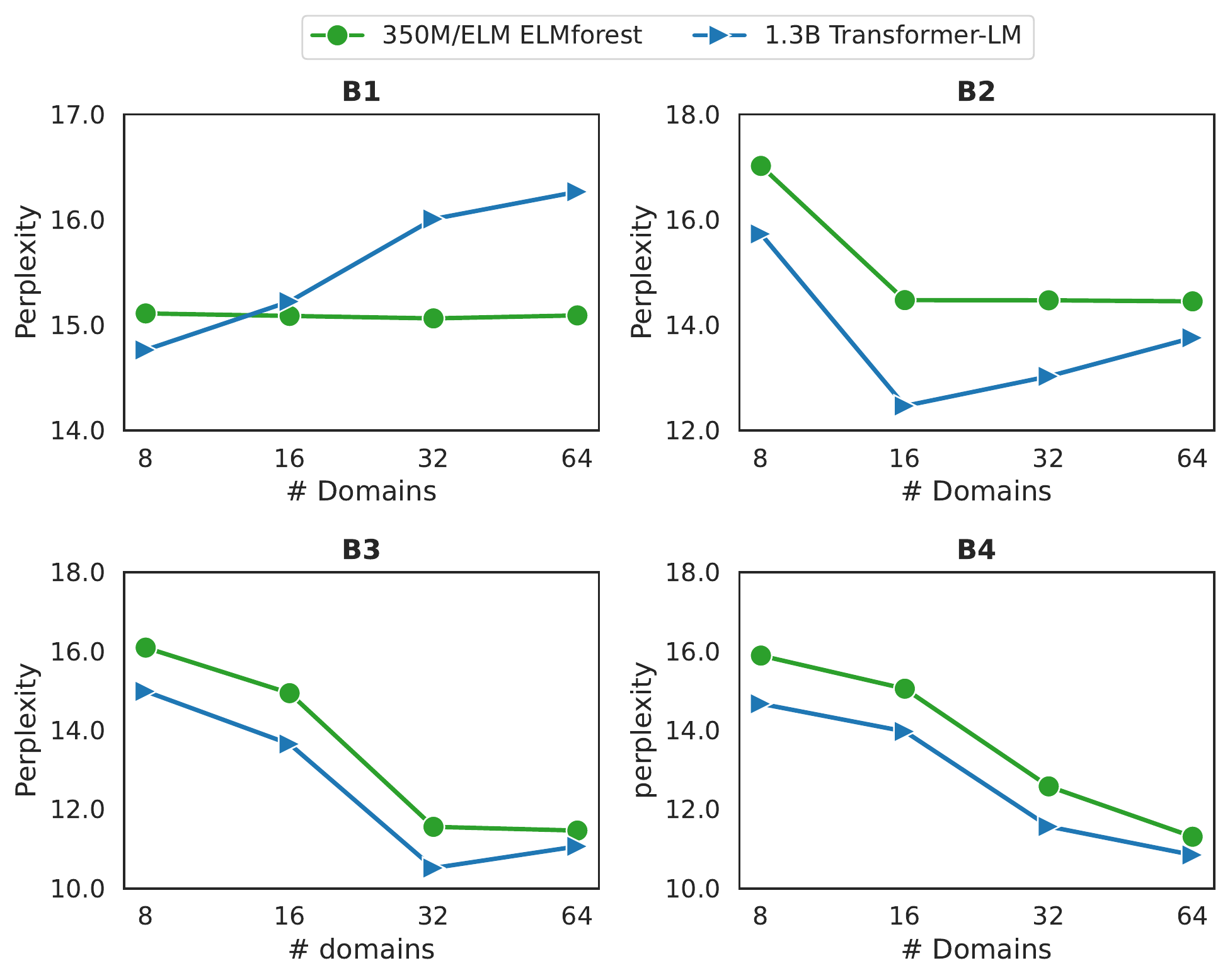}
    \caption{\textbf{As the number of training domains increase, \delm{}s retain or improve performance, while compute-matched \denselm{}s degrade.} Average test set perplexity over each batch of training domains (B$_1$ - B$_4$). For the \denselm{} experiment, we train a new \denselm from scratch on 8 (B1), 16 (B1 + B2), 32 (B1 + B2 + B3), and 64 (B1 + B2 + B3 + B4) domains, for 6144 GPUs hours each. For the \delm{} experiment, we use \densetomod to train on the batches incrementally. We observe that \denselm{}s suffer as one trains on more domains while \delm{}s retain or improve performance as domains are added incrementally. This reflects the "curse of multilinguality" phenomenon discovered for multilingual transformer LMs \citep{https://doi.org/10.48550/arxiv.2205.06266}.}
    \label{fig:group_perplexities}
\end{figure}

%% file: tables/datasets.tex
\begin{table*}[h]
\centering
\small
\begin{tabular}{lllr}
\toprule
& \bf Domain & \bf Corpus &  \bf \# Train (Eval.) Tokens  \\
\midrule 


\parbox[t]{3pt}{\multirow{8}{*}{\rotatebox[origin=c]{90}{\bf {\textsc{Training}}}}} & \oneb 
& 30M NewsWire sentences \citep{chelba2014billion} 
& 700M (10M) \\
& \cs 
& 1.89M full-text CS papers from S2ORC \citep{lo-wang-2020-s2orc} 
& 4.5B  (10M)\\
& \legal
& 2.22M U.S. court opinions \citep{caselaw2018} 
& 10.5B (10M) \\
 & \med      
& 3.2M full-text medical papers from  S2ORC \citep{lo-wang-2020-s2orc}  
& 9.5B (10M)\\

& \textsc{WebText}$^\dagger$
& 8M Web documents \citep{Gokaslan2019OpenWeb} 
& 6.5B  (10M)\\

& \textsc{RealNews}$^\dagger$
& 35M articles from \realnews  \cite{Zellers2019DefendingAN}
& 15B (10M) \\
& \reddit
& Reddit comments from pushshift.io \citep{baumgartner2020pushshift} 
& 25B  (10M)\\

& \textsc{Reviews}$^\dagger$
& 30M Amazon product reviews \citep{ni-etal-2019-justifying} 
& 2.1B (10M) \\

\midrule
& & 
& \bf{Total} \hspace{2mm}  73.8B (80M) \\
\end{tabular}

\begin{tabular}{lllr}
\midrule
& \bf Domain & \bf  Corpus & \bf \# Train (Eval.) Tokens \\
\midrule

 \parbox[t]{3pt}{\multirow{8}{*}{\rotatebox[origin=c]{90}{\bf \textsc{Evaluation}}}} & \aclpapers & 1.5K NLP papers from ACL \citep{Dasigi2021ADO} & 1M (1M)  \\
 & \textsc{Breaking News}$^\dagger$ 
& 20K English news articles, scraped using \citep{newspaper3k}  & 11M (1M) \\

& \textsc{Contracts}$^\dagger$ & 500 commercial legal contracts \citep{hendrycks2021cuad} & 1.5M (1M) \\
& \cord 
& 400K COVID-19 research papers \citep{wang2020cord19} &  60M (10M) \\
& \textsc{Github}
& 230K public Github code  \citep{githubcode} & 200M (10M) \\
& \gutenberg      
& 3.2M copyright-expired books \citep{gutenberg} & 3B (10M) \\

& \textsc{Tweets}$^\dagger$ 
& 1M English tweets from 2013-2018 & 8M (1M)
\\

&  \textsc{Yelp Reviews}$^\dagger$ 
& 6M Yelp restaurant reviews \citep{yelpreviews} & 600M (10M)\\

\bottomrule
\end{tabular}

\caption{\textbf{Multi-domain data corpus used in \S\ref{sec:core_results} and \S\ref{sec:analysis}}. Details of this corpus, both training and evaluation domains, including the size of our training and evaluation (i.e. validation and test) data in whitespace-separated tokens. We borrow these datasets from \citet{demix}. $\dagger$ indicates datasets we de-identify with regexes in Table \ref{tab:anonymization}. \reddit was de-identified by \citet{xu2020recipes}; we use their version.  Meta researchers did not collect any of the Reddit or Twitter data and the data was not collected on behalf of Meta. }
\label{tab:demix_corpus}
\end{table*}

%% file: tables/scaling_datasets_appendix.tex
\begin{table*}[t]
\centering
\tiny
\begin{tabular}{llrr}
\toprule
& \bf Domain & \bf Corpus   & \bf Batch \\
\midrule 


\parbox[t]{3pt}{\multirow{64}{*}{\rotatebox[origin=c]{90}{\bf {\textsc{Training Domains}}}}}
& \oneb 
& NewsWire sentences \citep{chelba2014billion}  &  B1 \\
& \cs 
& Full-text CS papers from S2ORC \citep{lo-wang-2020-s2orc} &  B1 \\
& \legal
& U.S. court opinions \citep{caselaw2018}   &  B1 \\
 & \med      
& Full-text medical papers from  S2ORC \citep{lo-wang-2020-s2orc}  &  B1 \\

& \textsc{OpenWebText}$^\dagger$
& OpenWebText Corpus \citep{Gokaslan2019OpenWeb}  &  B1 \\

& \textsc{RealNews}$^\dagger$
& Realnews Corpus \cite{Zellers2019DefendingAN}  &  B1 \\
& \reddit
& Reddit comments from pushshift.io \citep{baumgartner2020pushshift}  &  B1 \\

& \textsc{Reviews}$^\dagger$
& Amazon product reviews \citep{ni-etal-2019-justifying}  &  B1 \\

&  \textsc{Psychology} 
& Full-text Psychology papers from  S2ORC \citep{lo-wang-2020-s2orc}  &  B2 \\

&  \textsc{Chemistry} 
& Full-text Chemistry papers from  S2ORC \citep{lo-wang-2020-s2orc}  &  B2 \\

&  \textsc{Economics} 
& Full-text Economics papers from  S2ORC \citep{lo-wang-2020-s2orc}  &  B2 \\

&  \textsc{Engineering} 
& Full-text Engineering papers from  S2ORC \citep{lo-wang-2020-s2orc}  &  B2 \\

&  \textsc{Materials} 
& Full-text Materials papers from  S2ORC \citep{lo-wang-2020-s2orc}  &  B2 \\

&  \textsc{Geology} 
& Full-text Geology papers from  S2ORC \citep{lo-wang-2020-s2orc}  &  B2 \\

&  \textsc{Sociology} 
& Full-text Sociology papers from  S2ORC \citep{lo-wang-2020-s2orc}  &  B2 \\

&  \textsc{Business} 
& Full-text Business papers from  S2ORC \citep{lo-wang-2020-s2orc}  &  B2 \\

&  \textsc{C4} 
& Colossal Cleaned Common Crawl snapshot \citep{2019t5}  &  B3 \\

&  \textsc{Wikipedia} 
& 2022.03.01 Wikipedia snapshot \citep{wikidump}  &  B3 \\

&  \textsc{Stackoverflow}$^\dagger$ 
& Stackoverflow posts from The Pile \citep{https://doi.org/10.48550/arxiv.2101.00027}  &  B3 \\

&  \textsc{Twitter}$^\dagger$ 
& English tweets from 2013-2018 \citep{twitterapi} &  B3 \\

&  \textsc{Biology} 
& Full-text Biology papers from  S2ORC \citep{lo-wang-2020-s2orc}  &  B3 \\

&  \textsc{JavaScript} 
& JavaScript code from Github [MIT, BSD, Apache 2.0] \citep{huggingface} &  B3 \\

&  \textsc{HTML} 
& HTML code from Github [MIT, BSD, Apache 2.0] \citep{huggingface}&  B3 \\

&  \textsc{Gutenberg} 
& Copyright-expired books \citep{gutenberg}   &  B3 \\

&  \textsc{Political Science} 
& Full-text Political Science papers from  S2ORC \citep{lo-wang-2020-s2orc}  &  B3 \\

&  \textsc{Environmental Science} 
& Full-text Environmental Science papers from  S2ORC \citep{lo-wang-2020-s2orc}  &  B3 \\

&  \textsc{Physics} 
& Full-text Physics Papers from  S2ORC \citep{lo-wang-2020-s2orc}  &  B3 \\

&  \textsc{Mathematics} 
& Full-text Mathematics papers from  S2ORC \citep{lo-wang-2020-s2orc}  &  B3 \\

&  \textsc{Java} 
& Java code from Github [MIT, BSD, Apache 2.0] \citep{huggingface} &  B3 \\
&  \textsc{C} 
& C code from Github [MIT, BSD, Apache 2.0] \citep{huggingface}&  B3 \\

&  \textsc{C++} 
& C++ code from Github [MIT, BSD, Apache 2.0] \citep{huggingface} &  B3 \\

&  \textsc{Geography} 
& Full-text Geography papers from  S2ORC \citep{lo-wang-2020-s2orc}   &  B3 \\

&  \textsc{stackexchange}$^\dagger$ 
&  Stackexchange posts from The Pile \citep{https://doi.org/10.48550/arxiv.2101.00027}   &  B4 \\

&  \textsc{Philosophy} 
& Full-text Philosophy papers from  S2ORC \citep{lo-wang-2020-s2orc}  &  B4 \\

&  \textsc{CORD19} 
& COVID-19 research papers \citep{wang2020cord19}  &  B4 \\

&  \textsc{History} 
& Full-text History papers from  S2ORC \citep{lo-wang-2020-s2orc}  &  B4 \\

&  \textsc{Books Reviews}$^\dagger$ 
& Book review subset of Amazon reviews \citep{ni-etal-2019-learning}   & B4 \\

&  \textsc{Art} 
&  Full-text Art papers from  S2ORC \citep{lo-wang-2020-s2orc} &  B4 \\

&  \textsc{Python} 
& Python code from Github [MIT, BSD, Apache 2.0] \citep{huggingface} &  B4 \\

&  \textsc{C\#} 
& C\# code from Github [MIT, BSD, Apache 2.0] \citep{huggingface}  &  B4 \\

&  \textsc{PHP} 
& PHP code from Github [MIT, BSD, Apache 2.0] \citep{huggingface}  &  B4 \\

&  \textsc{Markdown} 
& Markdown code from Github [MIT, BSD, Apache 2.0] \citep{huggingface} &  B4 \\

&  \textsc{Code Contests} 
& Programming challenge questions and answers generated by AlphaCode \citep{li2022competition}  &  B4 \\

&  \textsc{Movie and TV reviews}$^\dagger$ 
& Movie and TV review subset of Amazon reviews \citep{ni-etal-2019-learning}  &  B4 \\

&  \textsc{Supreme Court Opinions} 
& Supreme Court Opinions from the Pile \citep{https://doi.org/10.48550/arxiv.2101.00027}   &  B4 \\

&  \textsc{Hacker News}$^\dagger$ 
&  Hacker news comments from The Pile \citep{https://doi.org/10.48550/arxiv.2101.00027}  &  B4 \\

&  \textsc{2021 WMT Newscrawl} 
& 2021 newswire sentences \citep{barrault-etal-2019-findings}  &  B4 \\

&  \textsc{N/A Semantic Scholar} 
& Full-text  papers marked N/A from  S2ORC \citep{lo-wang-2020-s2orc}  &  B4 \\

&  \textsc{OpenSubtitles}
& Movie subtitles \citep{lison-tiedemann-2016-opensubtitles2016}  &  B4 \\

&  \textsc{STORIES} 
& Filtered "story-like" Common Crawl documents  \citep{https://doi.org/10.48550/arxiv.1806.02847}  &  B4 \\

&  \textsc{BookCorpus} 
& Self-published novels \citep{Zhu_2015_ICCV}  &  B4 \\

&  \textsc{Ruby}
& Ruby code from Github [MIT, BSD, Apache 2.0] \citep{huggingface} &  B4 \\

&  \textsc{Sports and Outdoor Reviews}$^\dagger$ 
& Sports and Outdoor Reviews subset of Amazon Reviews \citep{ni-etal-2019-learning}  &  B4 \\

&  \textsc{US Patent Office} 
& US Patents from The Pile \citep{https://doi.org/10.48550/arxiv.2101.00027}  &  B4 \\

&  \textsc{ACL papers} 
& Full-text ACL papers from  S2ORC \citep{lo-wang-2020-s2orc}  &  B4 \\

&  \textsc{Yelp Reviews}$^\dagger$ 
& 6M Yelp restaurant reviews \citep{yelpreviews}  &  B4 \\

&  \textsc{Deepmind Mathematics} 
& Synthetically-generated mathematical question answer pairs \citep{saxton2018analysing}   &  B4 \\

&  \textsc{Clothing Reviews}$^\dagger$ 
& Clothing Reviews subset of Amazon Reviews \citep{ni-etal-2019-learning}  &  B4 \\

&  \textsc{Home and Kitchen Reviews}$^\dagger$ 
& Home and Kitchen subset of Amazon reviews \citep{ni-etal-2019-learning}  &  B4 \\

&  \textsc{Gaming Reddit Comments} 
& Reddit comments from pushshift.io,  gaming-topic \citep{baumgartner2020pushshift}  &  B4 \\

&  \textsc{Electronic Reviews} 
& Electronic Reviews subset of Amazon Reviews \citep{yelpreviews}  &  B4 \\

&  \textsc{Sports Comments} 
& Reddit comments from pushshift.io,  sports-topic \citep{baumgartner2020pushshift}  &  B4 \\

&  \textsc{Go} 
& Go code from Github [MIT, BSD, Apache 2.0] \citep{huggingface} &  B4 \\

&  \textsc{CSS} 
& CSS code from Github [MIT, BSD, Apache 2.0] \citep{huggingface} &  B4 \\
\midrule
\end{tabular}

\caption{64 training domains that make up our multi-domain training corpus, with the batch that they appear in for our scaling study in \S\ref{sec:scaling_domains}. $\dagger$ indicates datasets we de-identify with regexes in Table~\ref{tab:anonymization}.  \reddit was de-identified by \citet{xu2020recipes}; we use their version.  Meta researchers did not collect any of the Reddit or Twitter data and the data was not collected on behalf of Meta. }
\label{tab:scaling_datasets_appendix}
\end{table*}

%% file: tables/scaling_novel_datasets_appendix.tex
\begin{table*}[t]
\centering
\tiny

\begin{tabular}{llr}
\toprule
& \bf Domain & \bf  Corpus\\
\midrule
 \parbox[t]{3pt}{\multirow{16}{*}{\rotatebox[origin=c]{90}{\bf \textsc{Evaluation Domains}}}} & \textsc{IMDB} & IMDB reviews \citep{maas-EtAl:2011:ACL-HLT2011}  \\

& \textsc{Legal Contracts} & 500 commercial legal contracts \citep{hendrycks2021cuad}  \\
& \textsc{cscareerquestions}$^\dagger$       
& Reddit comments from pushshift.io, restricted to /r/cscareerquestions \citep{baumgartner2020pushshift} \\

& \textsc{india}$^\dagger$ 
& Reddit comments from pushshift.io, restricted to /r/india \citep{baumgartner2020pushshift}
\\

&  \textsc{Enron}$^\dagger$ 
& 6M Yelp restaurant reviews \citep{yelpreviews}\\

&  \textsc{hiphopheads}$^\dagger$ 
& Reddit comments from pushshift.io, restricted to /r/hiphopheads \citep{baumgartner2020pushshift} \\

&  \textsc{Congressional Bills}
& Congressional bills from BillSum \citep{kornilova2019billsum} \\

&  \textsc{Ireland Speeches}
& Irish parliamentary speeches, 1919-2013 \citep{parliamentaryspeeches} \\

&  \textsc{SQL} 
& SQL code from Github [MIT, BSD, Apache 2.0] \citep{huggingface} \\

&  \textsc{Rust}
& Rust code from Github [MIT, BSD, Apache 2.0] \citep{huggingface} \\

&  \textsc{PERL}
& PERL code from Github [MIT, BSD, Apache 2.0] \citep{huggingface} \\

&  \textsc{TeX}
& TeX code from Github [MIT, BSD, Apache 2.0] \citep{huggingface} \\

&  \textsc{FORTRAN}
& FORTRAN code from Github [MIT, BSD, Apache 2.0] \citep{huggingface} \\

&  \textsc{COVID19 Tweets}$^\dagger$ 
& Tweets with \#COVID-19 hashtag \citep{twitterapi} \\

&  \textsc{TOEFL Exam Responses}
& TOEFL exam responses \citep{Blanchard2013TOEFL11AC} \\

&  \textsc{Breaking News}$^\dagger$ 
& 20K English news articles, scraped using Newspaper3K \citep{newspaper3k}  \\

\bottomrule
\end{tabular}

\caption{32  domains that make up our novel domain corpus, including the size of our training and evaluation (i.e. validation and test) data, in whitespace-separated tokens. We borrow these datasets from \citet{demix}. $\dagger$ indicates datasets we de-identify with regexes in Table~\ref{tab:anonymization}. Meta researchers did not collect any of the Reddit or Twitter data and the data was not collected on behalf of Meta. }
\label{tab:scaling_novel_datasets_appendix}
\end{table*}

%% file: tables/anonymization.tex
\begin{table*}[t]
\centering
\small
\begin{tabular}{lllr}
\toprule
& \bf Category & \bf Link to Regex & Dummy Token  \\
\midrule 
& Email & \url{https://regex101.com/r/ZqsF9x/1} & \texttt{<EMAIL>} \\
& DART & \url{https://regex101.com/r/0tQ6EN/1} & \texttt{<DART>} \\
& FB User ID & \url{https://regex101.com/r/GZl5EZ/1} & \texttt{<FB\_USERID>} \\
& Phone Number & \url{https://regex101.com/r/YrDpPD/1} & \texttt{<PHONE\_NUMBER>} \\
& Credit Card Number & \url{https://regex101.com/r/9NTO6W/1} & \texttt{<CREDIT\_CARD\_NUMBER>} \\
& Social Security Number & \url{https://regex101.com/r/V5GPNL/1} & \texttt{<SSN>} \\
& User handles & \url{https://regex101.com/r/vpey04/1} & \texttt{<USER>} \\
\bottomrule
\end{tabular}

\caption{De-identification schema. We de-identify text using the regexes provided in the above links for the categories listed.}
\label{tab:anonymization}
\end{table*}

%% file: tables/base_results_extended.tex
\begin{table}[t]\small
\centering
\begin{tabular}{rcccccc}
\toprule
\multicolumn{7}{l}{\bf 125M -- 16 GPUs -- 80k updates} \\
\cmidrule{1-7}
& \denselmshort & \denselmshort & \demix & \delm{} & \random\ & \delm{} \\
& & \unbalanced & & (random init) & \textsc{Ensemble} & (seed init) \\ 
& 125M & 125M & 512M & 1B & 1B & 1B \\
\cmidrule{2-7}
Train   & 19.8 & 20.7  & 17.7 & 18.0  & 23.0  & \bf 17.2 \\
Novel   & 25.6 & 26.4  & 23.1 & 24.1  & 26.0  & \bf 22.4 \\
All     & 22.7 & 23.5  & 20.4 & 21.0  & 24.7  & \bf 19.8 \\
\bottomrule
\end{tabular}

\begin{tabular}{rcccccc}
\toprule
\multicolumn{7}{l}{\bf 350M -- 32 GPUs -- 32k updates} \\\cmidrule{1-7}
& \denselmshort & \denselmshort & \demix & \delm{} & \random\ & \delm{} \\
& & \unbalanced & & (random init) & \textsc{Ensemble} & (seed init) \\ 
& 350M & 350M & 1.8B & 2.8B & 2.8B & 2.8B \\
\cmidrule{2-7}
Train   & 16.3  & 16.7  & 15.0  & 15.3  & 19.9  & \bf 14.7 \\
Novel   & 20.8  & 21.2  & 19.9  & 21.3  & 23.1  & \bf 18.6 \\
All     & 18.5  & 19.0  & 17.5  & 18.3  & 21.5  & \bf 16.7 \\
\bottomrule
\end{tabular}

\begin{tabular}{rcccccc}
\toprule
\multicolumn{7}{l}{\bf 750M -- 64 GPUs -- 24k updates} \\ \cmidrule{1-7}
& \denselmshort & \denselmshort & \demix & \delm{} & \random\ & \delm{} \\
& & \unbalanced & & (random init) & \textsc{Ensemble} & (seed init) \\ 
& 750M & 750M & 3.8B & 6B & 6B & 6B \\
\cmidrule{2-7}
Train   & 14.7  & 14.9  & 13.5  & 14.4  & 17.4  & \bf 13.4 \\
Novel   & 19.3  & 19.8  & 17.7  & 19.3  & 20.9  & \bf 16.7 \\
All     & 17.0  & 17.4  & 15.6  & 16.9  & 19.2  & \bf 15.0 \\
\bottomrule
\end{tabular}

\begin{tabular}{rcccccc}
\toprule
\multicolumn{5}{l}{\bf 1.3B -- 128 GPUs -- 12k updates} \\ \cmidrule{1-7}
& \denselmshort & \denselmshort & \demix & \delm{} & \random\ & \delm{} \\
& & \unbalanced & & (random init) & \textsc{Ensemble} & (seed init) \\ 
& 1.3B & 1.3B & 7B & 10.4B & 10.4B & 10.4B \\
\cmidrule{2-7}
Train   &  14.2 & 15.0  & 13.7  & 13.3  & 17.4  & \bf 13.0 \\
Novel   &  18.4 & 19.5  & 17.6  & 17.8  & 20.4  & \bf 16.3 \\
All     &  16.3 & 17.3  & 15.6  & 15.6  & 18.9  & \bf 14.6 \\
\bottomrule \\
\end{tabular}
\caption{\textbf{\delm{}s trained with \densetomod outperform all baselines and ensemble variations across multiple model scales}. Average test-set perplexity ($\downarrow$) for each model scale (125M, 350M, 750M, 1.3B parameters) across the 8 training, 8 novel, and all 16 domains described in \S\ref{subsection:core_experimental_setup}. Total compute budget (in update numbers) and GPU usage are shown for each model size, and total parameters are shown for each model type at each size. \denselm{}s (here, abbreviated to \denselmshort{}) trained without balancing between data domains performs worse than \denselmshort trained with data balancing; hence, we only compare against the balanced \denselmshort setting in \S\ref{sec:core_results}. For \delm{}, we show results with 50\% dense training. }
\label{tab:base_results_extended}
\end{table}